\documentclass[lettersize,journal]{IEEEtran}

\usepackage{amsmath,amsfonts,amssymb}
\usepackage{algorithmic}
\usepackage{algorithm}
\usepackage{array}
\usepackage{booktabs}
\usepackage{multirow}
\usepackage[caption=false,font=normalsize,labelfont=sf,textfont=sf]{subfig}
\usepackage{textcomp}
\usepackage{stfloats}
\usepackage{verbatim}
\usepackage{graphicx}
\usepackage{cite}
\usepackage{xcolor}
\usepackage{balance}

\usepackage{url}
\usepackage[hidelinks,breaklinks=true]{hyperref}

\graphicspath{{../}{../In_Context_Forcing_Supplementary_Material/}}

\begin{document}

\title{In-Context Forcing: Uncovering Context Effects in Autoregressive Video Diffusion}

\author{
\textdagger Lingxiao Yang,
\textdagger Liu Liu,
\textdagger Moran Li,
Han Feng,
Wenjian Cao,
Jiangning Zhang,
and Ye Shi
\thanks{\textdagger~L. Yang, L. Liu and M. Li contributed equally to this work. \emph{(Corresponding author: Ye Shi.)}}
\thanks{L. Yang, L. Liu, and Y. Shi are with the School of Information Science and Technology, ShanghaiTech University, Shanghai 201210, China (e-mail: yanglx23@shanghaitech.edu.cn; liuliu2025@shanghaitech.edu.cn; shiye@shanghaitech.edu.cn).}
\thanks{M. Li , W. Cao and H. Feng are with Tencent Youtu Lab (e-mail: moranli.aca@gmail.com; weijiancao@tencent.com; whuerfff@whu.edu.cn).}
\thanks{J. Zhang is with the College of Computer Science and Technology, Zhejiang University, Hangzhou 310027, China (e-mail: 186368@zju.edu.cn).}
}

\markboth{Preprint}%
{Yang \MakeLowercase{\textit{et al.}}: In-Context Forcing}

\maketitle

\begin{abstract}
Current few-step autoregressive video diffusion models depend on previous fully denoised clean frames as context for all denoising steps of the current frame. However, these clean frames leak excessive local details, which causes the model to take shortcuts, resulting in compromised temporal semantics and dynamics. Inspired by the perspective of diffusion as masking, we explore the impact of noisy contexts on few-step autoregressive generation. Yet, simply applying contexts with the same noise levels provides insufficient guidance, leading to poor temporal consistency. To resolve this dilemma, we introduce In-Context Forcing, a progressive autoregressive paradigm that utilizes contexts with decreasing noise levels. By applying less masking to distant frames and more masking to adjacent ones, this approach provides adaptive guidance, effectively ensuring both robust temporal consistency and high inter-frame dynamics. Furthermore, by decoupling the strict dependence on previous clean frames, our paradigm enables cross-frame parallel denoising, achieving substantial inference acceleration without sacrificing performance. Extensive experiments on VBench demonstrate that our method significantly outperforms state-of-the-art approaches in both visual fidelity and inference speed.
\end{abstract}


\section{Introduction}
\IEEEPARstart{R}{ecent} advances in video generation~\cite{blattmann2023svd,blattmann2023align,brooks2024sora,gupta2024photorealistic,hacohen2024ltx,ho2022imagen,ho2022video,kong2024hunyuanvideo,polyak2024moviegen,villegas2022phenaki,wang2025wan,yang2025cogvideox} have enabled the synthesis of high-fidelity clips with remarkable temporal coherence and visual detail. Yet, most state-of-the-art approaches, typically based on diffusion models~\cite{ho2020ddpm,sohl2015nonequilibrium,song2021score}, rely on bidirectional attention mechanisms in Diffusion Transformers (DiTs)~\cite{peebles2023dit,wang2025wan} to generate entire sequences simultaneously. These methods often require multiple denoising steps, resulting in slow inference and limiting their use to offline scenarios. In contrast, many real-world interactive applications, such as game simulation~\cite{decart2024oasis,valevski2025diffusion,yu2025gamefactory}, live content creation~\cite{chen2024streaming,liang2025looking}, and robotics~\cite{li2025unified,yang2024learning}, require videos to be generated sequentially under strict real-time constraints. While the autoregressive generation paradigm naturally fits this streaming setup, AR-only models~\cite{bruce2024genie,kondratyuk2024videopoet,loong2024,weissenborn2020scaling,yan2021videogpt} often struggle to match the visual quality of their diffusion-based counterparts. Bridging this gap to enable sequential generation with both high fidelity and low latency is therefore a critical challenge for real-time applications. 

\begin{figure*}[!t]
\centering
\includegraphics[width=\textwidth]{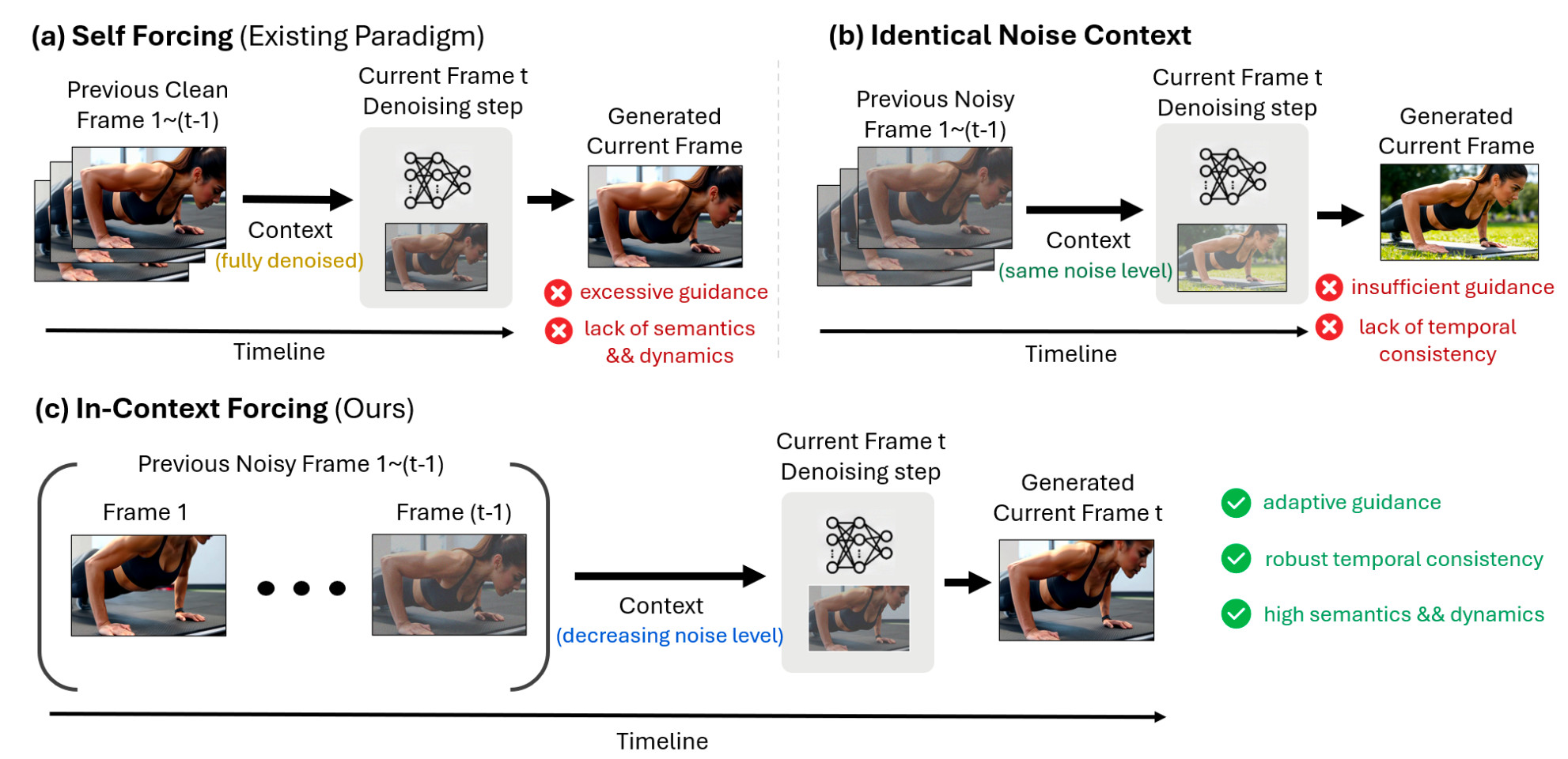}
\caption{ Comparison of contextual paradigms in few-step autoregressive video generation. 
(a) \textbf{Self Forcing}: depends on previous fully denoised clean frames as context, which causes excessive local detail leakage and compromises temporal semantics and dynamics.
(b) \textbf{Identical Noise Context}: applies contexts with the same noise level as the current frame, which masks excessive information and provides insufficient guidance.
(c) \textbf{In-Context Forcing}: introduces a progressive paradigm utilizing contexts with decreasing noise levels. By applying less masking to distant frames and more masking to adjacent ones, it provides adaptive guidance, ensuring robust temporal consistency and high inter-frame dynamics.\textit{Video frames are generated by a text-to-video (T2V) model.}}
\label{fig:context_paradigms}
\end{figure*}

To enable real-time video generation, recent studies have explored combining autoregressive paradigms with diffusion distillation. CausVid~\cite{yin2025causvid} introduces an asymmetric distillation strategy within the DMD~\cite{yin2024dmd} framework and follows the training paradigm of Diffusion Forcing~\cite{chen2024diffusionforcing,chen2025skyreels,sandai2025magi,chen2025historyguided}. However, this creates a critical train-test gap: training on ground-truth contexts while inferring with imperfect self-generated contexts leads to accumulated errors. Recently, Self Forcing~\cite{huang2025selfforcing} alleviates this train-test gap by reusing the previously generated clean frames as context to generate the next frame during training, as shown in Fig.~\ref{fig:context_paradigms}(a). However, these models depend on previous clean frames as context for all denoising steps of the current frame. These clean contexts leak local details, leading the model to directly replicate patterns from previous frames rather than treating them as meaningful guidance. Particularly during early denoising stages, this flaw causes the model to take shortcuts, as evidenced by the excessively high attention allocated to adjacent regions in previous frames, resulting in compromised temporal semantics and dynamics. Moreover, Rolling Forcing~\cite{liu2025rolling} adopts bidirectional attention within denoising windows to suppress error accumulation. However, this design inherently forces a split training paradigm: during training, a random exit flag determines which window writes the generator output, causing different frames within the same video to be predicted at heterogeneous denoising steps. Consequently, the resulting $\hat{x}_0$ estimates exhibit drastically different quality—some frames are coarse one-step predictions from pure noise while others are refined predictions near $t_0$—creating a mixed-quality output tensor that the DMD critic evaluates on. This trains the fake score to track a heterogeneous mixture distribution rather than the inference-time distribution, where every frame originates from $t_0$, fundamentally misaligning the optimization objective.

Inspired by the perspective of diffusion as masking, we explore the impact of noisy contexts to mitigate this issue. We first discover that simply applying contexts with the same noise levels matching the current frame masks excessive information, providing insufficient guidance and leading to poor temporal consistency, as illustrated in Fig.~\ref{fig:context_paradigms}(b). To resolve this dilemma, we introduce In-Context Forcing, a progressive autoregressive paradigm that utilizes contexts with decreasing noise levels. By applying less masking to distant frames and more masking to adjacent ones, this approach provides adaptive guidance, effectively ensuring both robust temporal consistency and high inter-frame dynamics.

While this progressive autoregressive paradigm offers significant advantages, maintaining distinct progressive contexts for each denoising step introduces additional complexity during training and inference. To address this during training, we propose a Step-wise Rolling Key-Value (KV) Cache mechanism. At each denoising step, this approach stores the KV cache corresponding to the current noise level. Then, it performs a bottom-up rolling update after each frame's denoising is completed, dynamically maintaining contexts at decreasing noise levels. Simultaneously, this mechanism accurately simulates inference behavior, successfully preserving train-test consistency while supporting our progressive autoregressive paradigm.

During inference, applying this context update strategy would incur substantial additional VRAM overhead due to the introduction of multi-level KV caches. Fortunately, by decoupling the strict dependence on fully denoised previous frames inherent in current few-step autoregressive models, our paradigm naturally supports inter-frame parallelization. Therefore, we introduce cross-frame causal attention, which not only reduces the memory requirement to a single KV cache but also achieves substantial inference acceleration on a single GPU through cross-frame parallel denoising.

In summary, our contributions are threefold.
\begin{itemize}
    \item We propose \textbf{In-Context Forcing}, a novel progressive autoregressive paradigm that mitigates excessive local detail leakage from clean contexts in current models. Our approach utilizes contexts with decreasing noise levels, specifically applying higher noise to adjacent frames and lower noise to distant ones. This provides adaptive guidance, effectively ensuring robust temporal consistency and high inter-frame dynamics.
    \item We introduce a \textbf{Step-wise Rolling KV Cache} to address the complexity of maintaining distinct progressive contexts during training. By storing the KV cache corresponding to the current noise level at each denoising step and performing a bottom-up rolling update after each frame's denoising is completed, this approach dynamically maintains contexts at decreasing noise levels.
    \item Extensive experiments on both standard short-video and extended long-video generation tasks demonstrate the superiority of our method. By decoupling the strict dependence on fully denoised previous frames, our paradigm naturally supports inter-frame parallelization. This enables the use of cross-frame causal attention during inference, which reduces the memory requirement to a single KV cache and achieves substantial acceleration without compromising performance.
\end{itemize}

\section{Related Work}

\subsection{Video Generation Models}
Video generation models~\cite{blattmann2023svd,kong2024hunyuanvideo,polyak2024moviegen,wang2025wan,yang2025cogvideox} typically employ bidirectional attention to generate all frames simultaneously, yielding high visual fidelity and temporal consistency. To support streaming video applications, recent autoregressive models~\cite{bruce2024genie,kondratyuk2024videopoet,ren2025nextblock,yan2021videogpt} equipped with causal attention adopt a next-token prediction paradigm. While this sequential framework reduces inference latency, it often suffers from degraded visual quality due to the accumulation of prediction errors. AR-diffusion hybrid models~\cite{arriola2025block,chen2024diffusionforcing,jin2025pyramid,liu2024vectorized,xie2025progressive} combine the advantages of both paradigms and show promising potential.

Among these methods, Diffusion Forcing~\cite{chen2024diffusionforcing} adds independent noise to each frame during training, enabling the model to adapt to different noise contexts during inference and support a pyramid-style inference schedule. However, its reliance on ground-truth training contexts creates an exposure bias that is significantly amplified during few-step inference. To address this, our In-Context Forcing proposes a train-test consistent formulation. By constructing training contexts via self-simulation rather than ground truth, we force the model to explicitly learn error correction, effectively bridging the train-test gap for robust progressive generation.

\subsection{Diffusion Distillation}
Video diffusion models produce high-quality results but suffer from slow iterative inference. To address this, distilling multi-step teachers into fewer-step students has emerged as a key direction to maintain generation quality while significantly reducing latency. Trajectory distillation and score distillation are the two main paradigms for this task. Trajectory distillation, such as consistency models~\cite{song2023consistency}, uses fewer-step student models to simulate the multi-step teacher's ODE trajectory, whereas score distillation matches output distributions across noise levels using the teacher's score function. Methods such as DMD~\cite{yin2024dmd} and SiD~\cite{zhou2024sid} minimize distribution divergence, e.g., KL or Fisher divergence, for high-fidelity one-step generation, while DMD2 incorporates self-simulation and gradient truncation to enable efficient few-step generation.

Score distillation particularly enables asymmetric setups, such as distilling a bidirectional teacher into a causal student. CausVid~\cite{yin2025causvid} first formalized this approach but suffers from a train-test gap due to its dependence on ground-truth noisy inputs during training. Self Forcing~\cite{huang2025selfforcing} mitigates this issue through self-simulation, reusing its previously generated clean frames as context to generate the next frame. Furthermore, Rolling Forcing~\cite{liu2025rolling} incorporates extrapolation techniques such as attention sinks and window-wise bidirectional attention, effectively alleviating quality drift over extended sequences to enable minute-level video generation.

To avoid the detail leakage caused by clean contexts in existing methods, our progressive autoregressive paradigm utilizes contexts with decreasing noise levels. This provides two key benefits. First, it offers adaptive guidance for robust temporal consistency and dynamics. Second, decoupling the reliance on fully denoised frames enables seamless integration with cross-frame causal attention, achieving memory-efficient parallel denoising and substantial inference acceleration.

\section{Methodology: In-Context Forcing}

We first formalize the autoregressive video diffusion paradigm in Section~\ref{sec:preliminaries}. We then introduce In-Context Forcing, a progressive paradigm that utilizes contexts with decreasing noise levels in Section~\ref{sec:icf}. In Section~\ref{sec:rolling_kv}, we describe the Step-wise Rolling KV Cache, a mechanism designed to maintain these distinct progressive contexts while ensuring train-test consistency. Finally, Section~\ref{sec:parallel} details the cross-frame parallel denoising scheme enabled by our paradigm, which achieves substantial inference acceleration.

\subsection{Preliminaries: Few-step Autoregressive Video Diffusion}
\label{sec:preliminaries}

The autoregressive video diffusion model combines the advantages of both diffusion and autoregressive models, preserving the autoregressive properties between frames while maintaining the iterative characteristics within each frame. Specifically, given a video consisting of $N$ frames $x^{1:N}=(x^1,x^2,\ldots,x^N)$, the frame-wise joint distribution can be expressed as
\begin{equation}
p(x^{1:N})=\prod_{i=1}^{N}p(x^i\mid x^{<i}),
\label{eq:ar_distribution}
\end{equation}
where $x^{<i}=(x^1,x^2,\ldots,x^{i-1})$ denotes the sequence of preceding $i-1$ frames. In each frame, given the $(T+1)$-step denoising schedule $\{t_0,t_1,\ldots,t_T\}$ where $t_0=0$ and $t_T=1000$, the step-wise conditional distribution of frame $i$ can be expressed as
\begin{equation}
p(x^i_0\mid x^{<i}_0)=\prod_{j=1}^{T-1}p(x^i_{t_{j-1}}\mid x^i_{t_j},x^{<i}_0),
\label{eq:step_distribution}
\end{equation}
where a diffusion process can be applied to achieve better visual quality. We can also generate a chunk of frames simultaneously to improve parallelism and frame-wise consistency. For simplicity of notation, each such chunk is treated as a single frame throughout this section.

Most existing few-step autoregressive video diffusion models are distilled from bidirectional teacher diffusion models using Distribution Matching Distillation (DMD) loss:
\begin{align}
\nabla \mathcal{L}_{\text{DMD}}
&= \mathbb{E}_{t}\left(\nabla_{\theta}\mathrm{KL}(p_{\text{fake},t}\Vert p_{\text{real},t})\right) \nonumber\\
&= -\mathbb{E}_{t,z,x_t}
\left(s_{\text{real}}(x_t,t)-s_{\text{fake}}(x_t,t)\right)
\frac{dG_{\theta}(z)}{d\theta},
\label{eq:dmd_loss}
\end{align}
where $z\sim\mathcal{N}(0,I)$ is the input noise, $G_{\theta}$ denotes the student model, and
\begin{equation}
x_t=\Psi(\hat{x}_0,\epsilon,t)=\alpha_t\hat{x}_0+\sigma_t\epsilon
\label{eq:noising}
\end{equation}
represents the noisy version of the student's predicted clean video $\hat{x}_0=G_{\theta}(z)$. Here, $s_{\text{fake}}$ and $s_{\text{real}}$ correspond to the score functions belonging to the student's and teacher's distributions, respectively. Since the DMD loss depends only on the student's distribution and imposes no constraints on the student architecture, it naturally supports asymmetric distillation, enabling knowledge transfer from a multi-step bidirectional teacher to a few-step autoregressive student.

\begin{figure*}[!t]
\centering
\includegraphics[width=\textwidth]{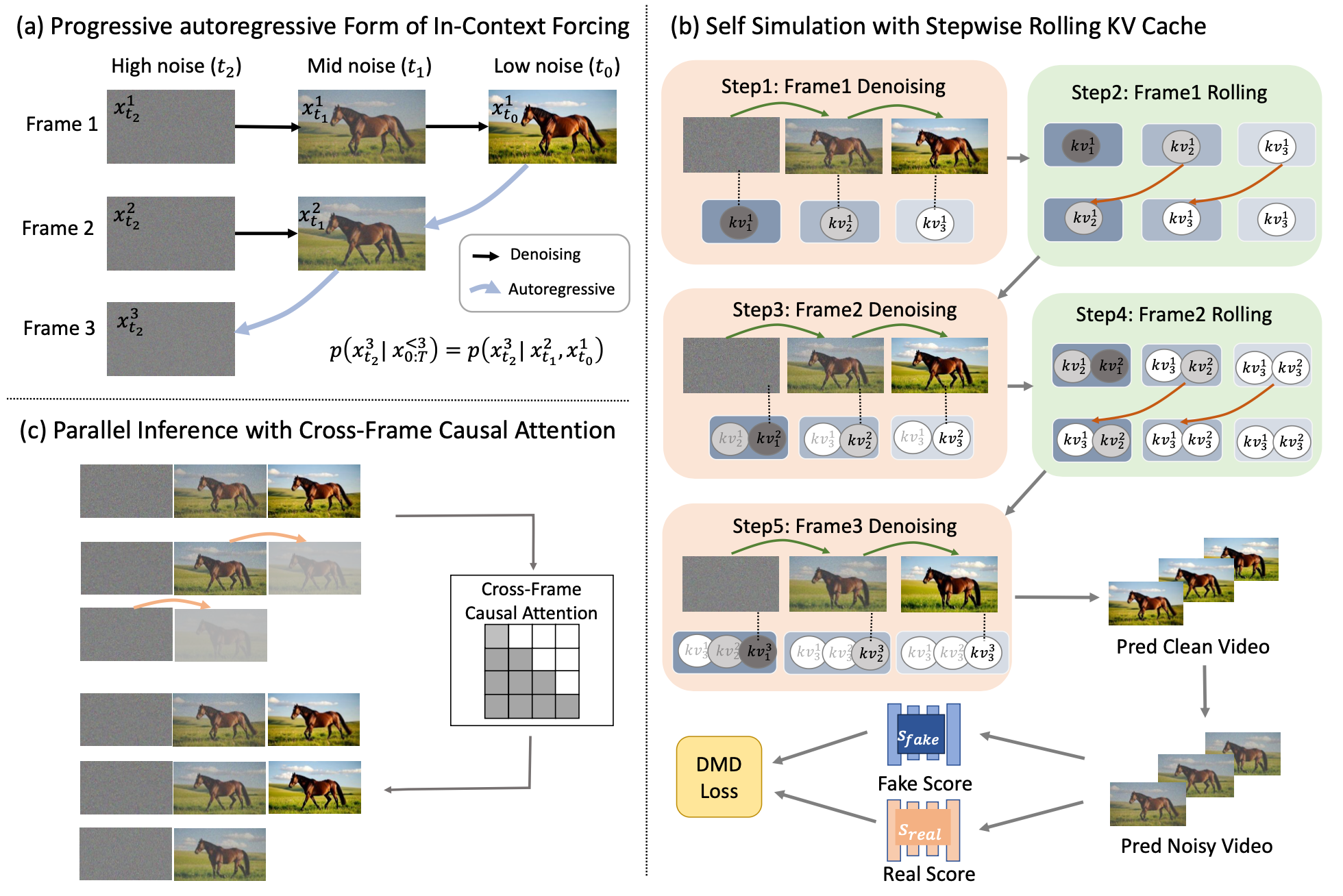}
\caption{ Overview of the proposed In-Context Forcing framework.
(a) \textbf{Progressive autoregressive form}: applies decreasing noise levels to preceding frames to provide adaptive guidance.
(b) \textbf{Step-wise Rolling KV Cache}: maintains these progressive contexts during training via self-simulation to ensure train-test consistency.
(c) \textbf{Parallel inference via cross-frame causal attention}: enables inter-frame parallel denoising to significantly accelerate inference.}
\label{fig:framework}
\end{figure*}

\subsection{In-Context Forcing}
\label{sec:icf}

While previous few-step autoregressive video diffusion models demonstrate strong performance, they typically treat inter-frame autoregression and intra-frame denoising as independent processes, which leads to two primary challenges. First, these models depend on previous clean frames as context for all denoising steps of the current frame. These clean contexts leak local details, causing the model to directly replicate patterns rather than treating them as meaningful guidance. Particularly during early denoising stages, this flaw leads the model to take shortcuts, which is evidenced by the high attention allocated to adjacent regions in previous frames. This results in compromised temporal semantics and dynamics. Second, this paradigm imposes a strict sequential dependency where subsequent frames can only begin denoising after the preceding ones are fully completed, thereby preventing cross-frame parallelization.

Building on this foundation and inspired by the perspective of diffusion as masking, we introduce the progressive autoregressive paradigm. Under this view, the noise level acts as a mask that hides local details. This allows us to provide adaptive guidance by explicitly controlling the noise levels of preceding frames. The conditional distribution of this paradigm can be expressed as
\begin{equation}
p(x^i_{t_j}\mid x^{<i}_{0:T})
=
p\left(
x^i_{t_j}
\mid
x^{i-1}_{t_{j-1}},\ldots,x^{i-k}_{t_{j-k}},\ldots
\right),
\label{eq:ours_ar}
\end{equation}
where $0<k\leq i$, and the noise level $t_{j-k}$ is clipped to $t_0$ when $j-k<0$. As shown in Fig.~\ref{fig:framework}(a), this design incorporates contextual information from preceding frames across decreasing noise levels to facilitate the denoising of the current frame. By explicitly applying higher noise to adjacent frames and lower noise to distant ones, this approach provides adaptive guidance. This effectively prevents the model from taking shortcuts, ensuring robust temporal consistency and high inter-frame dynamics. The underlying intuition is that early denoising stages primarily require coarse structural guidance to establish global layout and motion trajectories, whereas later stages demand fine-grained spatial details for precise texture synthesis. By masking adjacent frames heavily at early steps, our paradigm suppresses the leakage of low-level details that would otherwise be directly copied, forcing the model to rely on higher-level semantic signals from noisier contexts. Conversely, as denoising progresses and the current frame's content becomes more determined, progressively cleaner contexts from distant frames provide the precise spatial correspondence needed for coherent detail refinement. This coarse-to-fine contextual scheduling thus naturally aligns the information content of each context frame with the denoising objective at each step.

\subsection{In-Context Forcing Training with Step-wise Rolling KV Cache}
\label{sec:rolling_kv}

\textbf{Step-wise Rolling KV Cache.}
To maintain distinct progressive contexts for each denoising step, we propose the Step-wise Rolling KV Cache, a set of dedicated caches, each corresponding to a specific denoising timestep $t$. This design enables a progressive autoregressive schedule through recursive updates between caches, as shown in Fig.~\ref{fig:framework}(b). Specifically, during the denoising of a single frame, we store the KV cache corresponding to the noise level of each timestep $t$. After completing the denoising of a frame, a bottom-up rolling update is performed across adjacent caches to maintain the progressive causal context.

\textbf{Bridging the Train-Test Gap via Self-Simulation.}
In DMD-based few-step autoregressive diffusion, train-test consistency is particularly important because the critic directly optimizes the distribution of the student's predicted clean outputs. Rolling Forcing's split training violates this consistency: a random exit flag selects one window for the DMD loss, but different blocks within the selected window may correspond to different denoising stages. As a result, the resulting $\hat{x}_0$ tensor contains predictions with heterogeneous quality, ranging from near-clean estimates at $t_0$ to coarse one-step estimates from high-noise inputs. The fake score network is therefore trained to fit a mixed-quality output distribution, rather than the inference-time distribution in which each generated frame is obtained after the full denoising trajectory and output at $t_0$. This mismatch causes the DMD objective to optimize a biased target and can be repeatedly propagated across rolling windows, especially in long-video generation.

To achieve full train-test consistency, we integrate inference-stage scheduling into training via a self-simulation mechanism. Our KV cache design integrates seamlessly into this framework, requiring minimal modifications to Self Forcing while improving overall performance. Specifically, for a model with $T+1$ steps, we randomly sample a step $s\leq T$ during training and simulate step-wise rolling KV caches for $s+1$ steps. The resulting iterative output $X_{\theta}$ is then used as the student's prediction for computing the DMD loss.

However, directly computing parameter gradients along the entire diffusion trajectory often leads to prohibitive memory overhead. Therefore, we retain gradients only in the final denoising stage and truncate gradients flowing through the KV caches of preceding frames. Since we do not compute gradients for the KV caches, this truncation strategy also allows all KV caches except those needed for the current denoising step to be offloaded to the CPU, enabling memory consumption comparable to that of Self Forcing during training. Furthermore, even without CPU offloading, maintaining multiple KV caches increases the overall memory footprint by less than 30\% compared to the standard training baseline.

\begin{algorithm}[!t]
\caption{In-Context Forcing Training via Step-wise Rolling KV Cache}
\label{alg:training}
\begin{algorithmic}[1]
\STATE \textbf{Require:} Denoising timesteps $\{t_0,\ldots,t_T\}$
\STATE \textbf{Require:} Number of video frames $N$
\STATE \textbf{Require:} AR diffusion model $G_\theta$ which returns predicted clean image $\hat{x}^i_0$ and KV embeddings $kv^i_j$ of frame $i$ and timestep index $j$
\STATE \textbf{loop}
\STATE Initialize model output $X_\theta\leftarrow [\ ]$ and \textcolor{blue}{$\mathrm{KV}$ cache pool $\mathrm{KVPool}\leftarrow [[\ ]]$}
\STATE Sample $s\sim \mathrm{Uniform}(1,2,\ldots,T)$
\FOR{$i=1,\ldots,N$}
    \STATE Initialize $x^i_{t_T}\sim\mathcal{N}(0,I)$
    \FOR{$j=T,\ldots,s$}
        \STATE \textcolor{blue}{\textbf{KV}$\leftarrow \mathrm{KVPool}[j]$}
        \IF{$j=s$}
            \STATE Enable gradient computation
            \STATE Set $\hat{x}^i_0,kv^i_j\leftarrow G_\theta(x^i_{t_j};t_j,KV)$
            \STATE $X_\theta.\mathrm{append}(\hat{x}^i_0)$
            \STATE \textcolor{blue}{\textbf{KV}$.\mathrm{append}(kv^i_j)$}
            \STATE Disable gradient computation
            \STATE \textcolor{blue}{\textbf{KV}$\leftarrow \mathrm{KVPool}[s-1]$}
            \STATE Cache $kv^i_0\leftarrow G^{KV}_\theta(\hat{x}^i_0;0,KV)$
            \STATE $KV.\mathrm{append}(kv^i_0)$
        \ELSE
            \STATE Disable gradient computation
            \STATE Cache $kv^i_j\leftarrow G^{KV}_\theta(x^i_{t_j};t_j,KV)$
            \STATE \textcolor{blue}{\textbf{KV}$.\mathrm{append}(kv^i_j)$}
            \STATE Sample $\epsilon\sim\mathcal{N}(0,I)$
            \STATE Set $x^i_{t_{j-1}}\leftarrow \Psi(\hat{x}^i_0,\epsilon,t_{j-1})$
        \ENDIF
    \ENDFOR
    \FOR{$m=T,\ldots,s$}
        \STATE \textcolor{blue}{Set $\mathrm{KVPool}[m]\leftarrow \mathrm{KVPool}[m-1]$}
    \ENDFOR
\ENDFOR
\STATE Update $\theta$ via distribution matching loss
\STATE \textbf{end loop}
\end{algorithmic}
\end{algorithm}

\subsection{Parallel Denoising via Cross-frame Causal Attention}
\label{sec:parallel}

\textbf{Reducing Memory Footprint by a Unified KV Cache.}
While Section~\ref{sec:rolling_kv} introduces a progressive autoregressive framework that ensures train-test consistency, its inference-time memory overhead remains challenging. Maintaining multiple KV caches consumes substantial GPU memory, and offloading them to the CPU introduces communication latency that degrades inference speed. However, we observe that this progressive autoregressive paradigm, by decoupling the strict inter-frame dependencies, enables cross-frame parallelism, as shown in Fig.~\ref{fig:framework}(c). This allows us to replace multiple per-noise-level KV caches with a single unified cache, matching Self Forcing's memory footprint while avoiding CPU offloading overhead.

\begin{figure}[!t]
\centering
\includegraphics[width=\columnwidth]{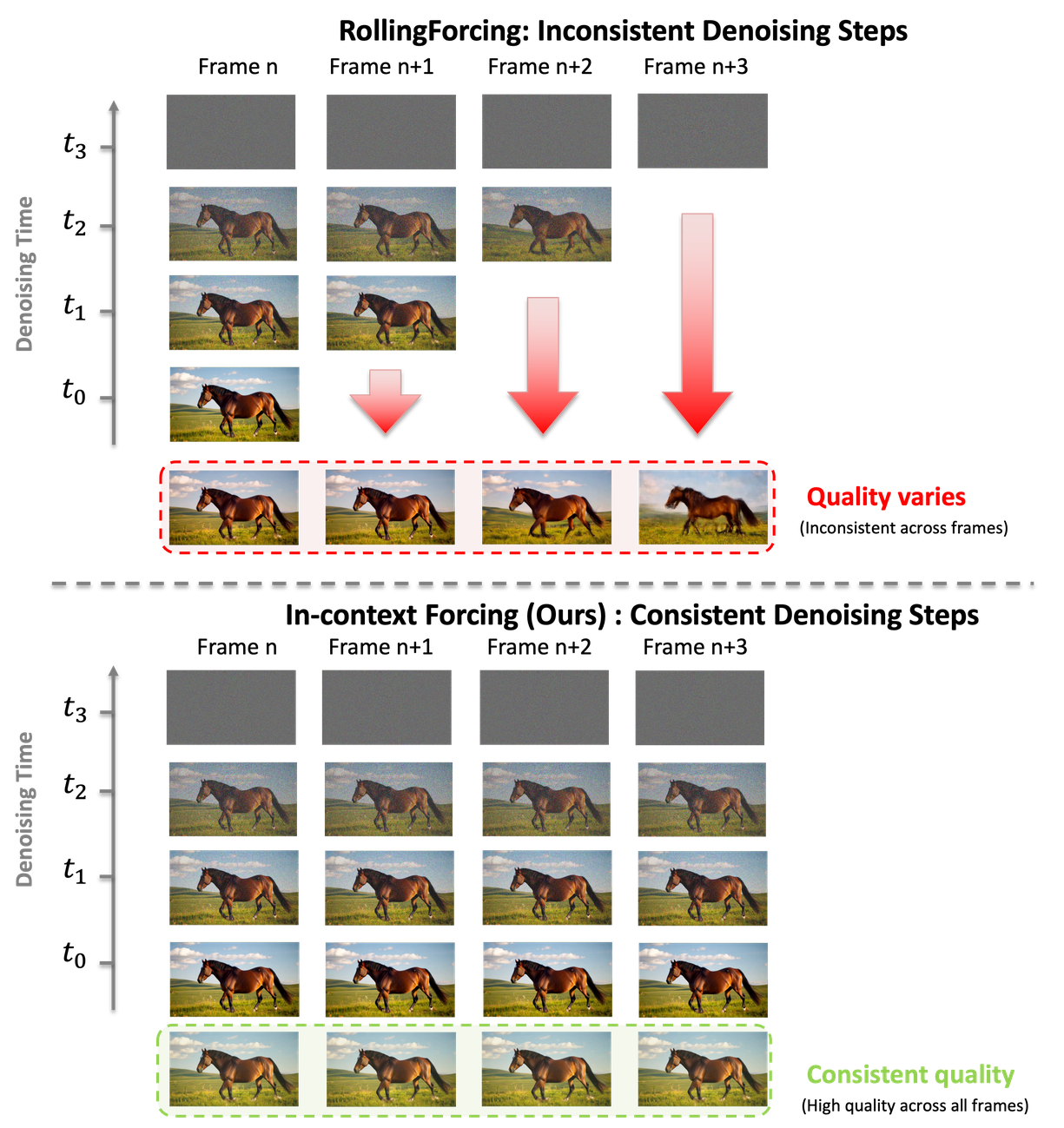}
\caption{\textbf{Visual Evidence of Train-Test Inconsistency: $\hat{x}_0$ Quality Disparity at Output.} \emph{Top:} Rolling Forcing's split training with random exit flag produces frames at heterogeneous denoising timesteps. At the output layer ($t_0$ row, red dashed box), the resulting $\hat{x}_0$ estimates exhibit drastically inconsistent quality---Frame~$n{+}1$ is a clean prediction (denoised from $t_0$), Frame~$n$ shows weak residual noise (from $t_1$), Frame~$n{+}2$ retains strong noise artifacts (from $t_2$), and Frame~$n{+}3$ remains near-pure noise (from $t_3$). This mixed-quality tensor corrupts DMD training by forcing the fake score to fit a heterogeneous mixture distribution. \emph{Bottom:} In-Context Forcing ensures every frame follows an identical denoising trajectory from $t_3 \to t_0$, producing uniformly high-quality $\hat{x}_0$ outputs across all frames (green dashed box). This strict train-test consistency avoids the mixed-output distribution that degrades motion diversity in Rolling Forcing.}
\label{fig:train_test_gap}
\end{figure}

Formally, we define a schedule matrix $\mathbb{T}\in\mathbb{R}^{N\times M}$ as:
\begin{equation}
\mathbb{T}=
\begin{bmatrix}
t_T & t_T & t_T & \cdots & t_T \\
t_{T-1} & t_T & t_T & \cdots & t_T \\
t_{T-2} & t_{T-1} & t_T & \cdots & t_T \\
\vdots & \vdots & \vdots & \ddots & \vdots \\
t_0 & t_1 & t_2 & \cdots & t_T \\
0 & t_0 & t_1 & \cdots & t_{T-1} \\
\vdots & \vdots & \vdots & \cdots & t_1 \\
t_0 & t_0 & t_0 & \cdots & t_0
\end{bmatrix},
\label{eq:schedule_matrix}
\end{equation}
where each row represents the noise levels for an $N$-frame sequence, and each column tracks the evolution across $M$ denoising steps. This matrix exhibits a key structural property: each column follows a monotonically non-decreasing trajectory from $t_0$ to $t_T$, encoding the progressive nature of our context schedule. The lower-triangular dominance of the matrix ensures that at any denoising step, a frame can only attend to preceding frames at equal or higher noise levels, which is precisely the constraint imposed by causal attention combined with our progressive noise scheduling. Importantly, the staircase pattern does not correspond to a strictly sequential autoregressive process where frame $n{+}1$ must wait until frame $n$ is fully denoised. Instead, it encodes a relaxed causal schedule: a new frame can begin denoising once its predecessors have advanced to sufficiently informative context states, enabling multiple frames to be processed in parallel while preserving the causal conditioning order. At each step, we identify active denoising indices and corresponding noise levels through inter-row comparison, process frames in parallel via cross-frame causal attention, and then re-noise outputs for the next iteration.

\textbf{Cross-frame Parallelism for Fast Inference.}
This parallelization addresses a key limitation in Self Forcing's inference efficiency: while KV caching reduces the number of tokens in a single attention computation, it leads to suboptimal GPU utilization. This inefficiency explains why the standard frame-wise configuration, despite performing fewer total attention operations, results in slower inference than the chunk-wise setup. Our approach effectively mitigates this sequential bottleneck through cross-frame parallelism, which dramatically improves GPU utilization across all denoising stages without sacrificing generation quality. Consequently, our method achieves a maximum relative speedup of 82\%, equivalent to a 45.1\% reduction in total inference time, under the frame-wise setting. Furthermore, compared to Self Forcing, it still yields a 9.3\% time reduction even in the already optimized chunk-wise configuration.

\textbf{Plug-and-Play Capability.}
Notably, by directly integrating this scheduling into the weights of the Self Forcing baseline in a training-free manner, we observe a seamless improvement in generation quality. We attribute this to the model's initialization: Self Forcing inherits its weights from CausVid, which is initialized similarly to Diffusion Forcing and inherently learns to process contexts across diverse noise levels. Consequently, this robust foundation in generalized contextual awareness is well preserved even after subsequent distillation.

\begin{algorithm}[!t]
\caption{In-Context Forcing Inference via Cross-frame Causal Attention}
\label{alg:inference}
\begin{algorithmic}[1]
\STATE \textbf{Require:} Denoising timesteps $\{t_0,\ldots,t_T\}$
\STATE \textbf{Require:} Number of generated video frames $M$
\STATE \textbf{Require:} Number of parallel sampling steps $N$
\STATE \textbf{Require:} AR diffusion model $G_\theta$ which returns predicted clean image $\hat{x}_0^i$ and $KV$ embeddings $kv$ of frame $i$.
\STATE \textbf{Require:} Schedule matrix $\mathbb{T}\in\mathbb{R}^{N\times M}$ , with $\mathbb{T}[i,j]$ as the noise level $t_{i,j}$ at denoising step $i$ for frame $j$.
\STATE Initialize model output $X_\theta\leftarrow [\ ]$
\STATE Initialize KV cache $KV\leftarrow [\ ]$
\STATE Initialize $X_\theta\leftarrow x^{1:M}_{t_T}\sim\mathcal{N}(0,I)$
\FOR{$s=2,\ldots,N$}
    \STATE $\mathrm{ActiveMask}\leftarrow \mathbb{T}[i,:]-\mathbb{T}[i-1,:]$
    \STATE $\mathrm{ActiveIndices}\leftarrow \mathrm{find}(\mathrm{ActiveMask}>0)$
    \STATE $t_{\mathrm{current}}\leftarrow \mathbb{T}[i,\mathrm{ActiveIndices}]$
    \STATE $x_{\mathrm{active}}\leftarrow X_\theta[\mathrm{ActiveIndices}]$
    \STATE $\hat{x}_0,kv\leftarrow G_\theta(x_{\mathrm{active}};t_{\mathrm{current}},KV)$
    \STATE $KV.\mathrm{append}(kv)$
    \STATE $t_{\mathrm{next}}\leftarrow \mathbb{T}[i+1,\mathrm{ActiveIndices}]$
    \STATE Sample $\epsilon\sim\mathcal{N}(0,I)$
    \STATE $x_{\mathrm{next}}\leftarrow \Psi(\hat{x}_0,\epsilon,t_{\mathrm{next}})$
    \STATE $X_\theta[\mathrm{ActiveIndices}]\leftarrow x_{\mathrm{next}}$
\ENDFOR
\STATE \textbf{return} $X_\theta$
\end{algorithmic}
\end{algorithm}

\section{Experiments}

\subsection{Implementation Details}

\textbf{Training.}
We implement In-Context Forcing using the causal variant of the Wan2.1-T2V-1.3B~\cite{wang2025wan} architecture as the base model. The corresponding variant based on bidirectional attention can generate 5-second 480p video clips in 20 to 50 steps. We follow CausVid's initialization procedure~\cite{yin2025causvid}, in which the model is finetuned using 16K ODE pairs sampled from the original bidirectional model together with the causal attention mask. We conduct experiments using a 4-step chunk-wise autoregressive diffusion model, where each chunk jointly generates 3 frames unless otherwise specified. The prompts used for ODE initialization and DMD distillation are drawn exclusively from text prompts provided by a filtered and LLM-extended version of VidProM~\cite{wang2024vidprom}. Consequently, no video data is required for the experiments. For DMD asymmetric distillation, we employ the Wan2.1-T2V-14B model based on bidirectional attention as the teacher, serving as both the real and fake score networks. Additional hyperparameters are provided in Appendix~\ref{sec:supp_impl}.

\textbf{Evaluation.}
We holistically evaluate our method on both standard short-video and extended long-video generation tasks. For quantitative assessment, we adopt VBench~\cite{huang2024vbench} to measure comprehensive dimensions such as visual quality, temporal consistency, and semantic alignment. Specifically, we evaluate standard short clips alongside 30-second long sequences, for which we utilize inference-time extrapolation techniques following Rolling Forcing~\cite{liu2025rolling}. Furthermore, we conduct a rigorous blind, randomized user study to gauge overall human preference through A/B testing, recorded as Better, Same, or Bad. Finally, inference speed is measured by throughput, i.e., FPS, following Self Forcing~\cite{huang2025selfforcing}.

\subsection{Comparison with State-of-the-Art}

We compare our method with the most relevant open-source models. The baselines include two diffusion models, Wan2.1~\cite{wang2025wan} and LTX-Video~\cite{hacohen2024ltx}, where Wan2.1 also serves as the base model used in our distillation. We further include several autoregressive models such as MAGI-1~\cite{sandai2025magi}, SkyReels-V2~\cite{chen2025skyreels}, NOVA~\cite{deng2025nova}, Pyramid Flow~\cite{jin2025pyramid}, CausVid~\cite{yin2025causvid}, Self Forcing~\cite{huang2025selfforcing}, and Rolling Forcing~\cite{liu2025rolling}.

\begin{table*}[!t]
\caption{Comparison with state-of-the-art methods on short-video generation. We compare In-Context Forcing with current models in terms of inference speed and VBench scores on standard short clips.}
\label{tab:sota_short}
\centering
\begin{tabular}{lcccccc}
\toprule
\multirow{2}{*}{Model} & \multirow{2}{*}{\#Params} & \multirow{2}{*}{Throughput (FPS) $\uparrow$} & \multicolumn{4}{c}{Evaluation Scores $\uparrow$} \\
\cmidrule(lr){4-7}
& & & Total Score & Quality Score & Semantic Score & Dynamic Degree \\
\midrule
Wan2.1 1.3B~\cite{wang2025wan} & 1.3B & 0.78 & 84.26 & 85.30 & 80.09 & 61 \\
LTX-Video~\cite{hacohen2024ltx} & 1.9B & 8.98 & 80.00 & 82.30 & 70.79 & 46 \\
MAGI-1~\cite{sandai2025magi} & 4.5B & 0.19 & 79.18 & 82.04 & 67.74 & 42 \\
SkyReels-V2~\cite{chen2025skyreels} & 1.3B & 0.49 & 82.67 & 84.70 & 74.53 & 37 \\
NOVA~\cite{deng2025nova} & 0.6B & 0.88 & 80.12 & 80.39 & 79.05 & 46 \\
Pyramid Flow~\cite{jin2025pyramid} & 2B & 6.7 & 81.72 & 84.74 & 69.62 & 16 \\
CausVid~\cite{yin2025causvid} & 1.3B & 17.0 & 81.20 & 84.05 & 69.80 & 62 \\
Self Forcing*~\cite{huang2025selfforcing} & 1.3B & 17.0 & 83.95 & 84.67 & 81.05 & 63 \\
Rolling Forcing*~\cite{liu2025rolling} & 1.3B & 13.6 & 83.03 & 83.69 & 80.38 & 43 \\ 
\textbf{In-Context Forcing*} & 1.3B & \textbf{18.4} & \textbf{84.34} & \textbf{84.99} & \textbf{81.77} & \textbf{72} \\
\bottomrule
\end{tabular}
\vspace{2mm}

\footnotesize{*The evaluation was conducted with a consistent random seed across experiments.}
\end{table*}

\begin{figure*}[!t]
\centering
\includegraphics[width=\textwidth]{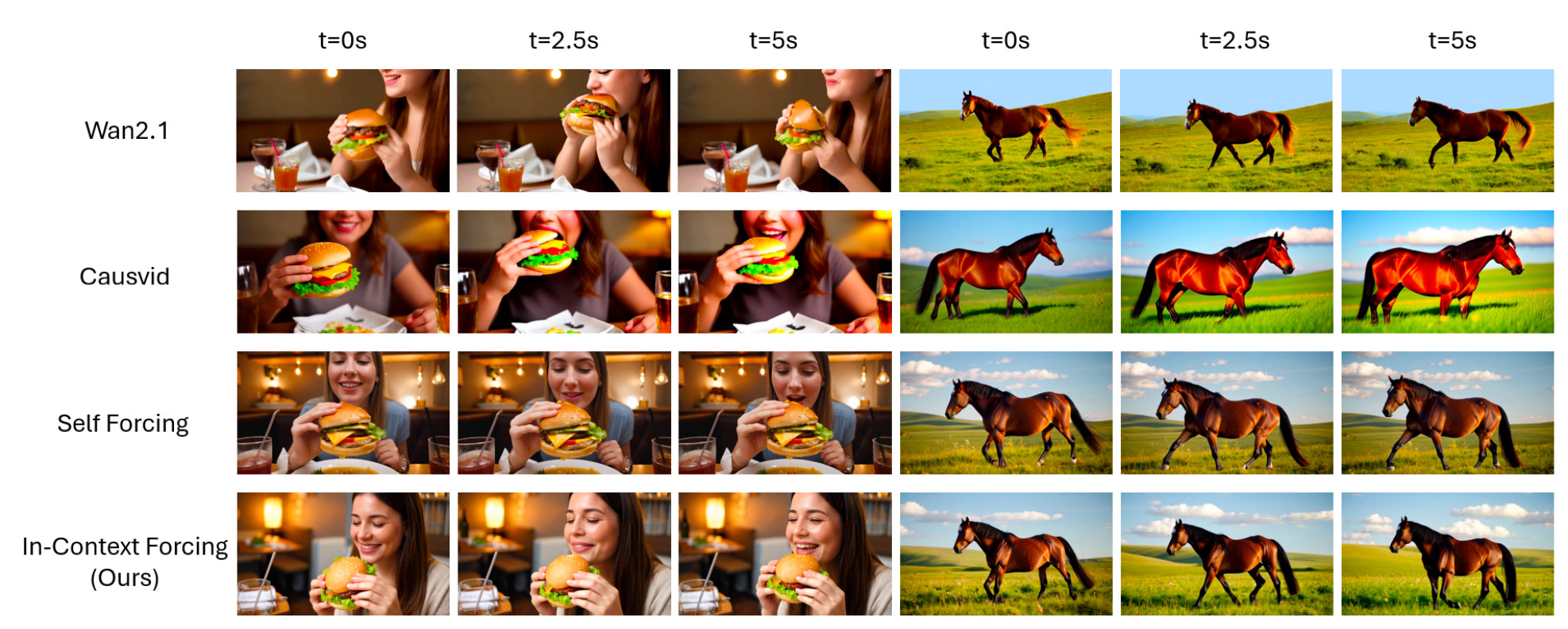}
\caption{Qualitative comparison with state-of-the-art methods of the same architecture, i.e., 1.3B parameters, on short-video generation. Our method produces frames with richer motion dynamics and stronger semantic coherence with the text prompt, while Self Forcing and CausVid tend to replicate patterns from preceding frames, resulting in static or repetitive motion. \textit{Video frames featuring a person are generated by a text-to-video (T2V) model.}}
\label{fig:short_qualitative}
\end{figure*}

\begin{figure}[!t]
\centering
\includegraphics[width=\columnwidth]{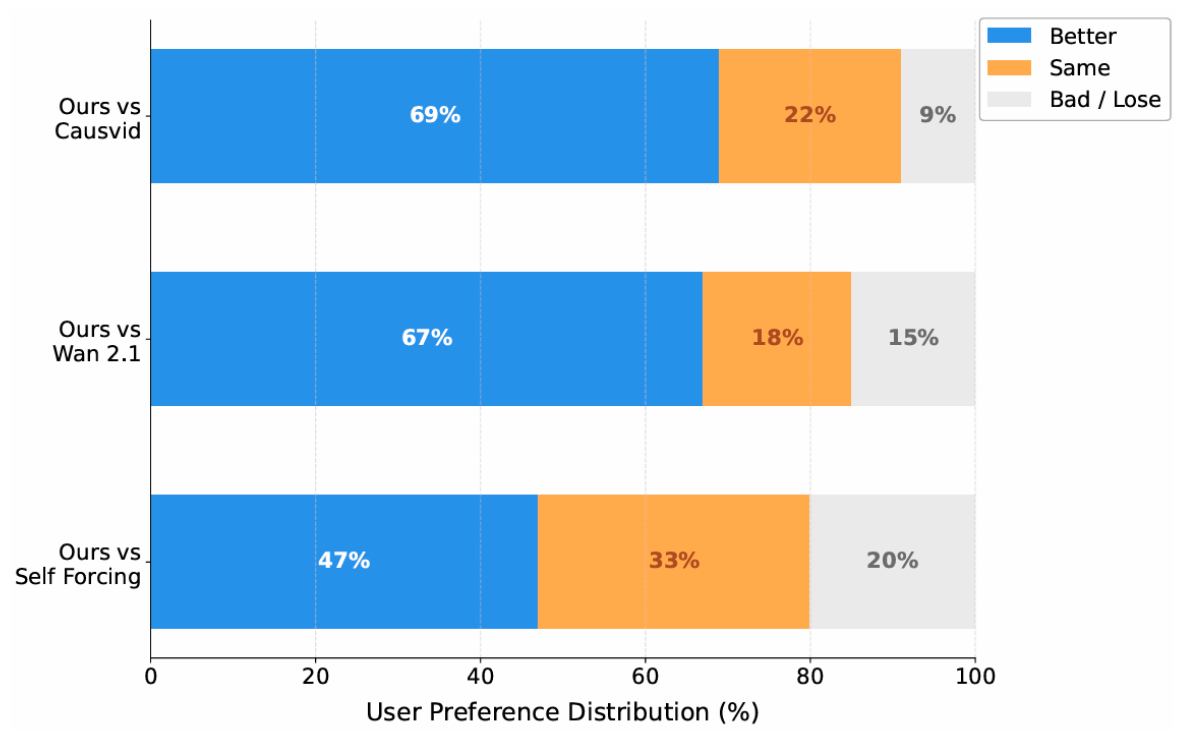}
\caption{User preference study on short-video generation.  Participants performed blind A/B testing between our method and each baseline, rating videos as ``Better,'' ``Same,'' or ``Worse.'' Our approach is consistently preferred across all comparisons.}
\label{fig:user_study}
\end{figure}

\begin{figure*}[!t]
\centering
\includegraphics[width=\textwidth]{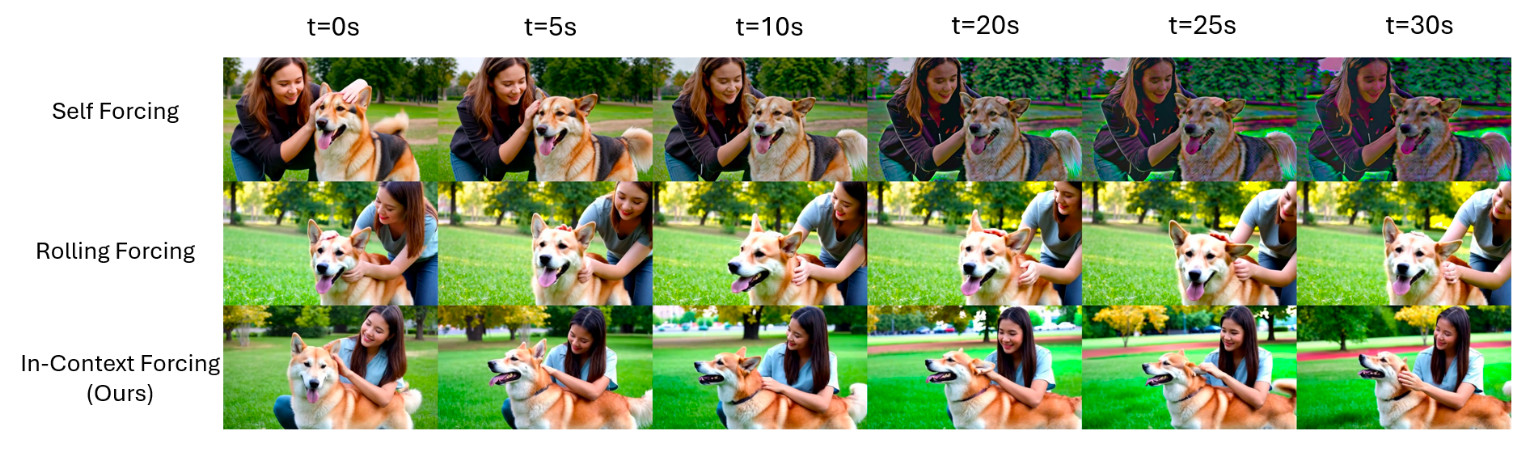}
\caption{ Qualitative comparison with state-of-the-art methods of the same architecture, i.e., 1.3B parameters, on 30-second long-video generation. Our method sustains diverse motion throughout extended sequences.\textit{Video frames are generated by a text-to-video (T2V) model.}}
\label{fig:long_qualitative}
\end{figure*}

\textbf{Superior Video Quality and Semantic Alignment.}
The qualitative and quantitative results for both standard short clips and long sequences are shown in Table~\ref{tab:sota_short}, Table~\ref{tab:long_video}, Fig.~\ref{fig:short_qualitative}, Fig.~\ref{fig:user_study}, and Fig.~\ref{fig:long_qualitative}. Our method surpasses all baseline models in VBench scores, demonstrating superior visual quality, dynamics, and semantic alignment across both short and long video generation tasks, as well as in the user preference study. This improvement stems from the diverse contextual signals used during both training and inference, which ensure that the full information from preceding frames influences the current frame only in the later denoising stages. This prevents the model from directly copying frame-specific patterns and thereby enhances temporal dynamics. Moreover, since the contextual features come from higher noise levels during the early denoising stages, they primarily convey coarse semantic structure rather than fine details. This provides stronger global semantic guidance, which in turn improves semantic coherence in the generated videos.

\textbf{User Study Setup.}
We conduct a blind, randomized A/B user study to assess human preference. From VBench~\cite{huang2024vbench}, we randomly sample a diverse set of extended prompts covering varied scenes, subjects, and motion types, and use them to generate videos with our method as well as three baselines: Wan2.1~\cite{wang2025wan}, CausVid~\cite{yin2025causvid}, and Self Forcing~\cite{huang2025selfforcing}. For each prompt and each baseline, the two videos are presented side by side in a randomized left-right order with method identities hidden. Participants are then asked to compare overall generation quality by selecting one of three labels: \emph{Better}, \emph{Same}, or \emph{Worse}. Each video pair is rated by multiple independent participants to mitigate individual bias, and the aggregated results across all comparisons are reported in Fig.~\ref{fig:user_study}. Further details are given in Appendix~\ref{sec:supp_user}.

\textbf{Accelerated Inference via Parallel Denoising.}
In addition, our method yields substantial improvements in inference speed, as detailed in Table~\ref{tab:inference_speed}. Under the frame-wise setting, where standard autoregressive models typically suffer from severe sequential bottlenecks and low GPU utilization, our cross-frame parallelism dramatically mitigates this issue. It boosts the throughput from 8.9 FPS to 16.2 FPS, achieving an 82\% relative speedup, or a 45.1\% reduction in total inference time, over Self Forcing. Furthermore, even under the already optimized chunk-wise setting, our model still achieves a 9.3\% reduction in total inference time. These gains stem from our cross-frame causal attention mechanism, which processes multiple frames simultaneously to maximize hardware utilization, resulting in significantly accelerated inference without loss of accuracy.

\begin{table*}[!t]
\caption{Inference speed comparison between Self-forcing and In-Context Forcing. The relative time reductions are highlighted in green.}
\label{tab:inference_speed}
\centering
\small
\setlength{\tabcolsep}{1.5pt}
\renewcommand{\arraystretch}{1.03}
\begin{tabular*}{0.7\textwidth}{@{\extracolsep{\fill}}lcccc@{}}
\toprule
Model &
\begin{tabular}[c]{@{}c@{}}Total Time $\downarrow$\end{tabular} &
\begin{tabular}[c]{@{}c@{}}Diffusion Time $\downarrow$\end{tabular} &
\begin{tabular}[c]{@{}c@{}}VAE Time $\downarrow$\end{tabular} &
FPS $\uparrow$ \\
\midrule
Self-forcing (frame-wise) & 9.10s & 7.25s & 1.85s & 8.9 \\
In-Context Forcing (frame-wise) & \textbf{5.00s} {\tiny (\textcolor{blue}{-45.1\%})} & \textbf{3.15s} {\tiny (\textcolor{blue}{-56.6\%})} & 1.85s & \textbf{16.2} \\
Self-forcing (chunk-wise) & 4.85s & 3.00s & 1.85s & 17.0 \\
In-Context Forcing (chunk-wise) & \textbf{4.40s} {\tiny (\textcolor{blue}{-9.3\%})} & \textbf{2.54s} {\tiny (\textcolor{blue}{-15.3\%})} & 1.85s & \textbf{18.4} \\
\bottomrule
\end{tabular*}
\vspace{-2mm}
\end{table*}

\begin{table}[!t]
\caption{VBench evaluation for 30-second long-video generation.}
\label{tab:long_video}
\centering
\small
\setlength{\tabcolsep}{2pt}
\begin{tabular}{lcccc}
\toprule
Method &
\begin{tabular}[c]{@{}c@{}}Total\\Score\end{tabular} &
\begin{tabular}[c]{@{}c@{}}Quality\\Score\end{tabular} &
\begin{tabular}[c]{@{}c@{}}Semantic\\Score\end{tabular} &
\begin{tabular}[c]{@{}c@{}}Dynamic\\Degree\end{tabular} \\
\midrule
Self Forcing & 80.44 & 80.95 & 78.40 & 52.51 \\
Rolling Forcing & 82.40 & 82.81 & 80.82 & 40.63 \\
\textbf{In-Context Forcing} & \textbf{82.97} & \textbf{83.49} & \textbf{80.90} & \textbf{86.40} \\
\bottomrule
\end{tabular}
\end{table}

\subsection{Ablation Study}

\begin{figure*}[!t]
\centering
\includegraphics[width=\textwidth]{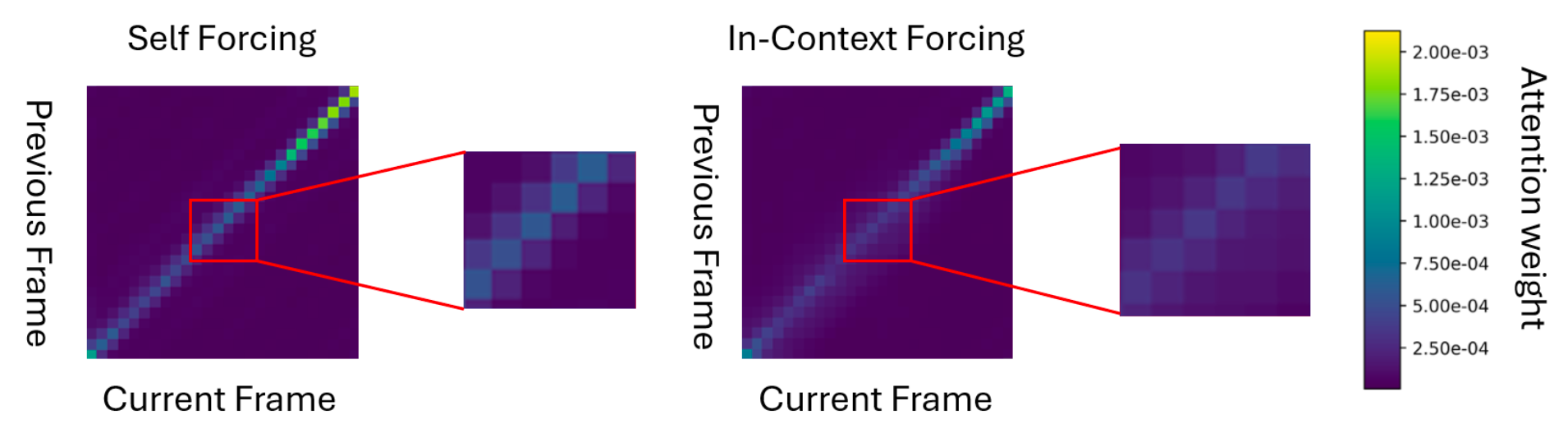}
\caption{ Cross-frame causal attention maps in the initial denoising step. Visualizing attention from the current frame's Query (Q) to the preceding frame's Key (K), Self Forcing exhibits a sharply concentrated diagonal that over-emphasizes local details. In contrast, our method yields a smoother, more dispersed distribution to maintain a broader semantic receptive field.}
\label{fig:attention_map}
\end{figure*}

\textbf{Attention Map Analysis.}
Fig.~\ref{fig:attention_map} visualizes the cross-frame causal attention during the initial denoising step. As highlighted in the zoomed-in regions, baselines such as Self Forcing exhibit a sharply concentrated diagonal, placing excessive attention on the exact spatial locations of the preceding frame. Consequently, this strict spatial localization causes the model to over-emphasize local details, leading to excessive pattern replication. In contrast, our In-Context Forcing yields a smoother, more dispersed attention distribution. By utilizing contexts with decreasing noise levels, it prevents local over-reliance and maintains a broader semantic receptive field. As generation progresses, our attention maps adaptively converge to those of Self Forcing to facilitate fine-grained detail refinement. A complete visualization together with its extraction protocol is provided in Appendix~\ref{sec:supp_attn}.

\textbf{Plug-and-Play Improvement.}
We compare the clean context used in Self Forcing with the progressive context introduced in our approach. Experiments show that even when the progressive context is applied in a plug-and-play manner to the same model weights trained with Self Forcing, it still achieves plug-and-play performance improvement, as shown in Table~\ref{tab:vbench_six}. Notably, without specific training on the progressive context, the model demonstrates strong generalization capability by effectively supporting progressive context inference, which may be attributed to the model initialization strategy described in Section~\ref{sec:parallel}.

\begin{table*}[!t]
\caption{Quantitative comparison with state-of-the-art methods across six dimensions of VBench.}
\label{tab:vbench_six}
\centering
\small
\begin{tabular}{lcccccc}
\toprule
Model & Dynamic Degree & Subject Consistency & Aesthetic Quality & Object Class & Spatial Relationship & Scene \\
\midrule
Self Forcing & 0.633 & 0.956 & 0.660 & 0.938 & 0.814 & 0.553 \\
In-Context Forcing* & 0.653 & \textbf{0.962} & 0.663 & 0.942 & 0.819 & 0.563 \\
\textbf{In-Context Forcing} & \textbf{0.719} & 0.959 & \textbf{0.665} & \textbf{0.964} & \textbf{0.842} & \textbf{0.572} \\
\bottomrule
\end{tabular}
\vspace{2mm}

\footnotesize{*Our method using the model weights from Self Forcing with the modified inference procedure in Section~\ref{sec:parallel}.}
\end{table*}

\textbf{Training for Semantic and Motion Enhancement.}
Table~\ref{tab:vbench_six} presents detailed VBench evaluations across six dimensions, including three visual quality metrics, i.e., dynamic degree, subject consistency, and aesthetic quality, and three semantic alignment metrics, i.e., object class, spatial relationship, and scene. Our method significantly outperforms Self Forcing in terms of dynamic degree. We attribute this improvement to the high-noise contexts applied during early denoising stages, which effectively prevent the direct replication of preceding frames and thereby enhance overall temporal dynamics. Furthermore, our approach demonstrates clear improvements in semantic alignment, particularly in the trained model. This suggests that the contextual paradigm of In-Context Forcing better captures global semantic information, serving as an additional benefit of utilizing high-noise contexts during the initial denoising phase. The gain is especially pronounced in long-video generation, where the dynamic degree gap over Rolling Forcing widens dramatically (86.40 vs.\ 40.63, Table~\ref{tab:long_video}). This confirms that Rolling Forcing's train-test inconsistency disproportionately degrades motion diversity over extended sequences, as the mismatch is repeatedly propagated across successive rolling windows. In contrast, our cross-frame causal attention preserves strict train-test consistency throughout generation, sustaining motion diversity regardless of video length.

\section{Conclusion}

This work identified a key limitation in few-step autoregressive video diffusion models: the isolation between inter-frame autoregression and intra-frame denoising. Strictly relying on fully denoised frames as context leaks excessive local details, causing shortcut learning that degrades temporal dynamics and prevents parallel generation. To solve these issues, we introduced In-Context Forcing, a progressive paradigm that utilizes contexts with decreasing noise levels. By applying higher noise to adjacent frames and lower noise to distant ones, this approach provides adaptive guidance to prevent pattern replication. To ensure strict train-test consistency, we further proposed the Step-wise Rolling KV Cache through self-simulation. Finally, by decoupling the reliance on clean contexts, our method enables cross-frame causal attention for inter-frame parallel denoising. Extensive evaluations have demonstrated that our approach achieves superior semantic alignment and dynamic degree, along with substantial inference acceleration. Limitations and future directions are discussed in Appendix~\ref{sec:supp_limit}.

\section{Acknowledgments}

This work was supported by the National Natural Science Foundation of China under Grants 62406195, the HPC Platform of ShanghaiTech University, and Key Laboratory of Intelligent Perception and Human-Machine Collaboration (ShanghaiTech University), Ministry of Education.

\bibliographystyle{IEEEtran}
\bibliography{refs}


\twocolumn[%
\begin{center}
\vskip0.2em
{\Huge Supplementary Material for In-Context Forcing: Uncovering Context Effects in Autoregressive Video Diffusion\par}
\vskip1.0em
\end{center}
\vspace{0.5em}
]

\appendices

\section*{Overview of the Appendices}
The appendices provide additional details and experimental results that complement the main paper. Appendix~\ref{sec:supp_video} describes the accompanying video demonstrations. Appendix~\ref{sec:supp_impl} reports additional implementation details. Appendix~\ref{sec:supp_exp} presents extended experimental analyses, including attention-map and denoising-procedure visualizations. Appendix~\ref{sec:supp_user} details the user study, covering both the A/B preference protocol and the comprehensive score evaluation. Appendix~\ref{sec:supp_limit} discusses limitations and future directions. Finally, Appendix~\ref{sec:supp_qual} provides additional qualitative results for short, 30-second, and 60-second generation. Figures and tables in the appendices continue the numbering of the main paper.

\section{Video Demonstrations}
\label{sec:supp_video}
\textbf{Supplementary Video Comparisons.}
While the static frames provided in the main paper and these appendices illustrate the visual quality of our generated samples, dynamic video demonstrations better convey temporal dynamics and motion smoothness. We therefore release side-by-side video comparisons between our In-Context Forcing and the Self Forcing baseline as part of the supplementary material accompanying this paper.

\section{Additional Implementation Details}
\label{sec:supp_impl}
We employ Wan2.1-T2V-14B as the teacher model and its causal variant Wan2.1-T2V-1.3B as the student model, with a classifier-free guidance (CFG) scale of $3$. For optimization, we use AdamW with $\beta_1=0$, $\beta_2=0.999$, and a weight decay of $0.01$. The learning rates for the student model $G_\theta$ and the fake score network $s_{\text{fake}}$ are set to $2\times10^{-6}$ and $4\times10^{-7}$, respectively, with an update ratio of $5$ (i.e., the student model updates once every $5$ updates of the fake score network). Training is conducted with a total batch size of $64$ for $1{,}900$ iterations, using the exponential moving average (EMA) with a decay rate of $0.99$. Consistent with the evaluation protocol of the base Wan2.1 model~\cite{wang2025wan} and Self Forcing~\cite{huang2025selfforcing}, we evaluate on VBench~\cite{huang2024vbench} by rewriting the test prompts using Qwen2.5-7B-Instruct~\cite{yang2024qwen25}.

\section{Additional Experimental Results}
\label{sec:supp_exp}

\subsection{Attention Map Visualization}
\label{sec:supp_attn}
\begin{figure*}[!t]
\centering
\includegraphics[width=\textwidth]{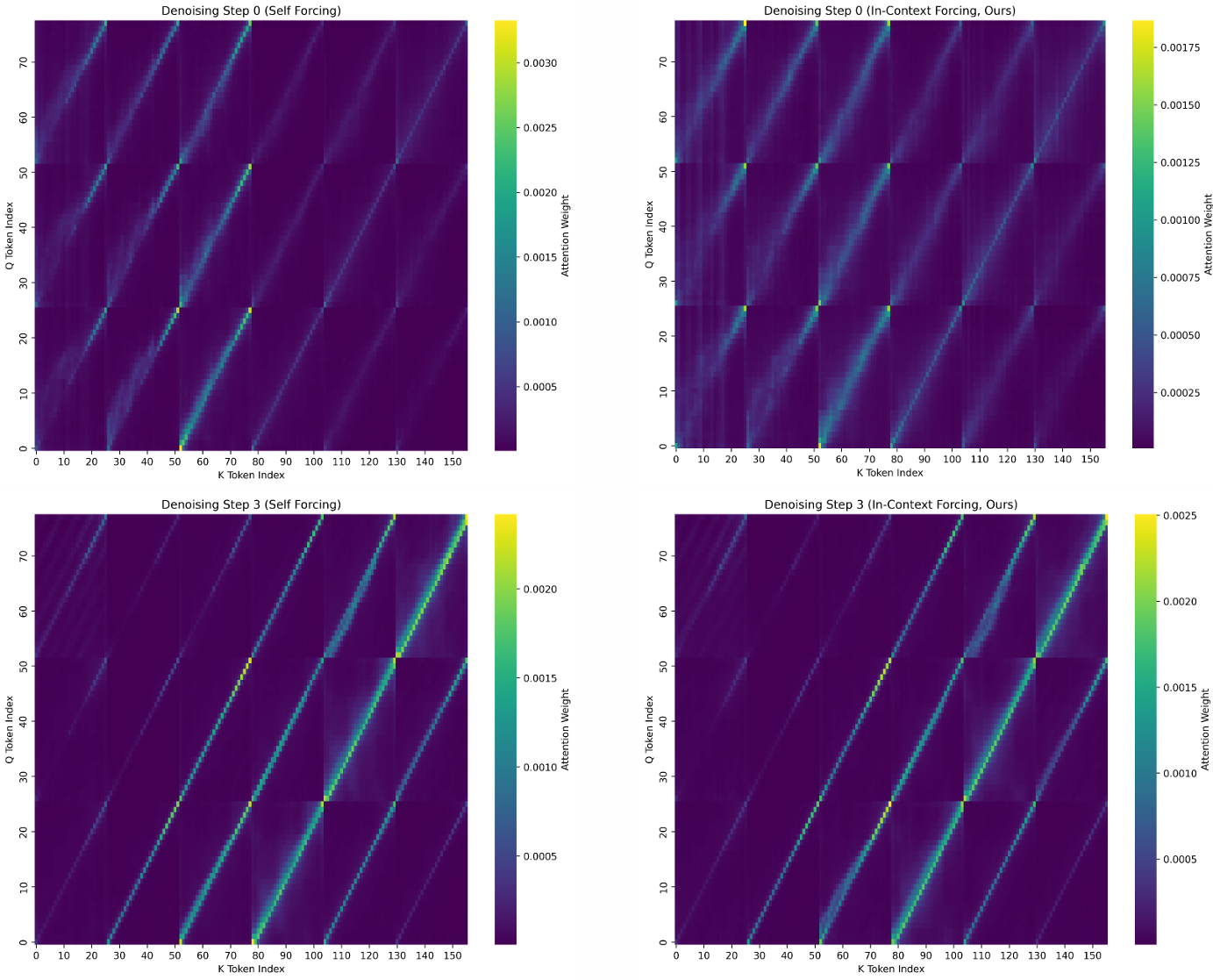}
\caption{\textbf{Attention Map Comparison between Self Forcing and In-Context Forcing.} The visualization depicts attention patterns during the second chunk generation (with token indices ranging from $0$ to $78$ for the previous chunk and $79$ to $156$ for the current chunk) under consistent prompts, transformer blocks, and random seeds. During early denoising at Step 0 (top row), Self Forcing rigidly concentrates attention on the exact spatial locations of previous frames. In contrast, In-Context Forcing effectively mitigates this over-reliance and prevents direct pattern replication by maintaining a more dispersed attention distribution with a stronger focus on the current frame. In late denoising at Step 3 (bottom row), both methods converge to a sharply concentrated diagonal, reflecting our approach's adaptive transition from global semantic alignment to fine-grained detail refinement.}
\label{fig:supp_attn}
\end{figure*}

\textbf{Implementation Details.}
We compare the contextual differences between Self Forcing and our In-Context Forcing through attention-map visualization in Fig.~\ref{fig:supp_attn}. The visualization focuses on the second chunk generation process, where key tokens $0$ to $78$ correspond to the previous chunk and $79$ to $156$ correspond to the current chunk. Each chunk contains $4680$ tokens, and with a downsample rate of $60$, each chunk is represented by $78$ tokens in the visualization. Notably, the attention visualization presented in Fig.~\ref{fig:attention_map} is explicitly extracted from this configuration. Specifically, it crops the region spanning Key token indices $52$ to $78$ and Query token indices $105$ to $131$, which corresponds to the attention map between the 5th frame and the 2nd frame. All maps are generated under identical prompts, denoising steps, transformer blocks, and random seeds, with average pooling applied to downsample high-dimensional attention maps and averaging across all attention heads for clarity. Frame boundaries are marked with grid lines, with the horizontal axis representing key length and the vertical axis indicating query length.

\textbf{Attention Maps in Early Denoising Stage.}
Distinct differences emerge in the early denoising stages: Self Forcing, which strictly relies on fully denoised contexts, demonstrates excessive pattern replication from previous frames. This over-reliance on local details is manifested as disproportionately high attention weights at exact corresponding spatial locations of preceding frames, which even surpass the attention allocated to the current frame. In contrast, our In-Context Forcing utilizes contexts with higher noise levels to provide adaptive guidance. This maintains a broader semantic receptive field during initial denoising, evidenced by a significantly more dispersed attention distribution. Rather than collapsing into sharp point-to-point mappings, the attention smoothly spans broader spatial neighborhoods while maintaining a stronger relative emphasis on the current frame.

\textbf{Attention Maps in Late Denoising Stage.}
As generation progresses to later denoising stages (e.g., Step 3), In-Context Forcing naturally incorporates contexts with lower noise levels. Consequently, its attention distribution transitions from the previously dispersed pattern to a sharply concentrated diagonal, adaptively converging to the localized attention patterns of Self Forcing. This structural convergence explicitly demonstrates our method's capability to shift its focus toward fine-grained detail refinement in the final stages, perfectly complementing the robust global semantic alignment established during the early denoising phases.

\subsection{Denoising Procedure Visualization}
\label{sec:supp_denoise}
\begin{figure*}[!t]
\centering
\includegraphics[width=\textwidth]{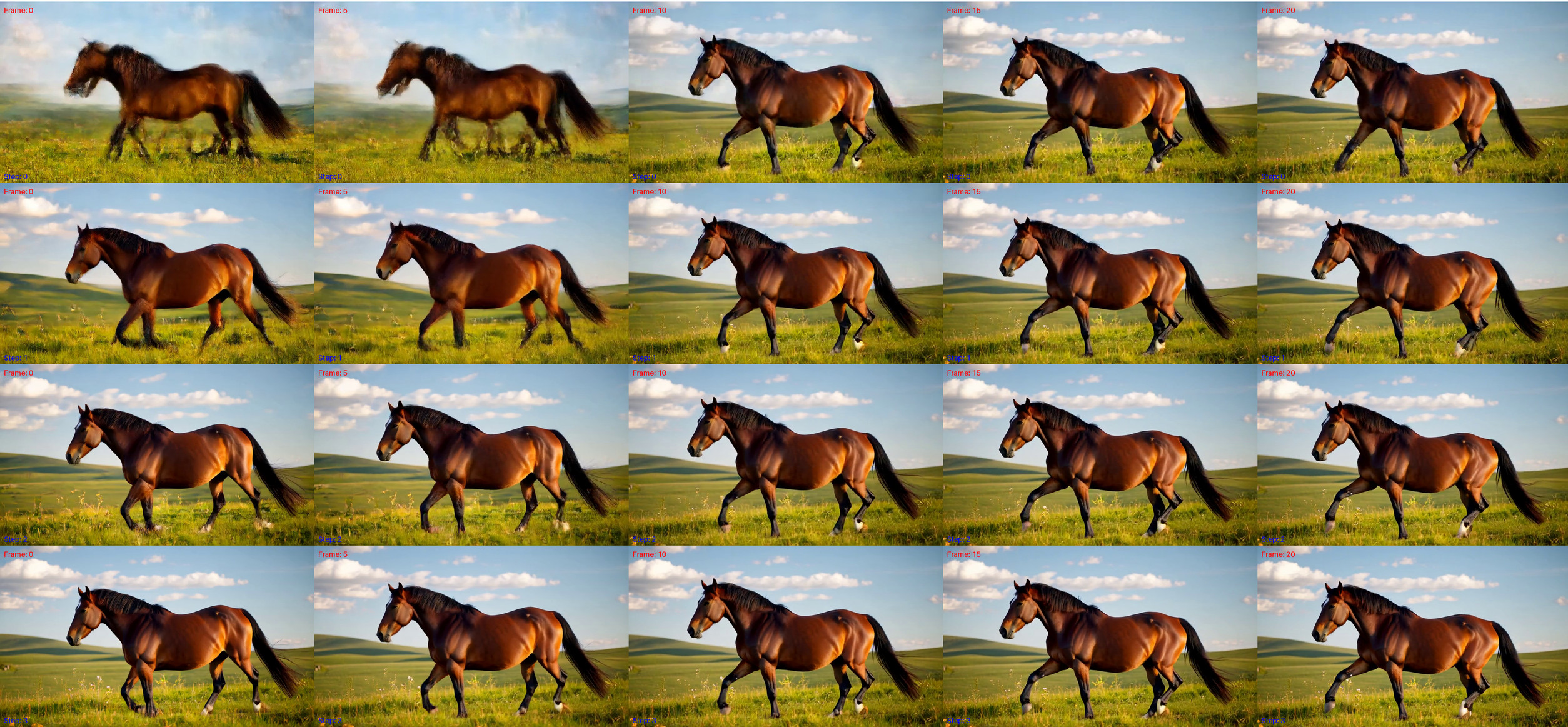}
\caption{\textbf{Self Forcing Denoising Visualization.} Results across sequential denoising steps (rows) for the first 20 frames sampled every 5 frames (columns). By strictly relying on fully denoised contexts, Self Forcing leads to an overly rapid detail restoration as early as the first denoising step. This over-reliance causes excessive local pattern replication, heavily compromising temporal dynamics and severely limiting the capacity for refinement in later stages.}
\label{fig:supp_denoise_sf}
\end{figure*}

\begin{figure*}[!t]
\centering
\includegraphics[width=\textwidth]{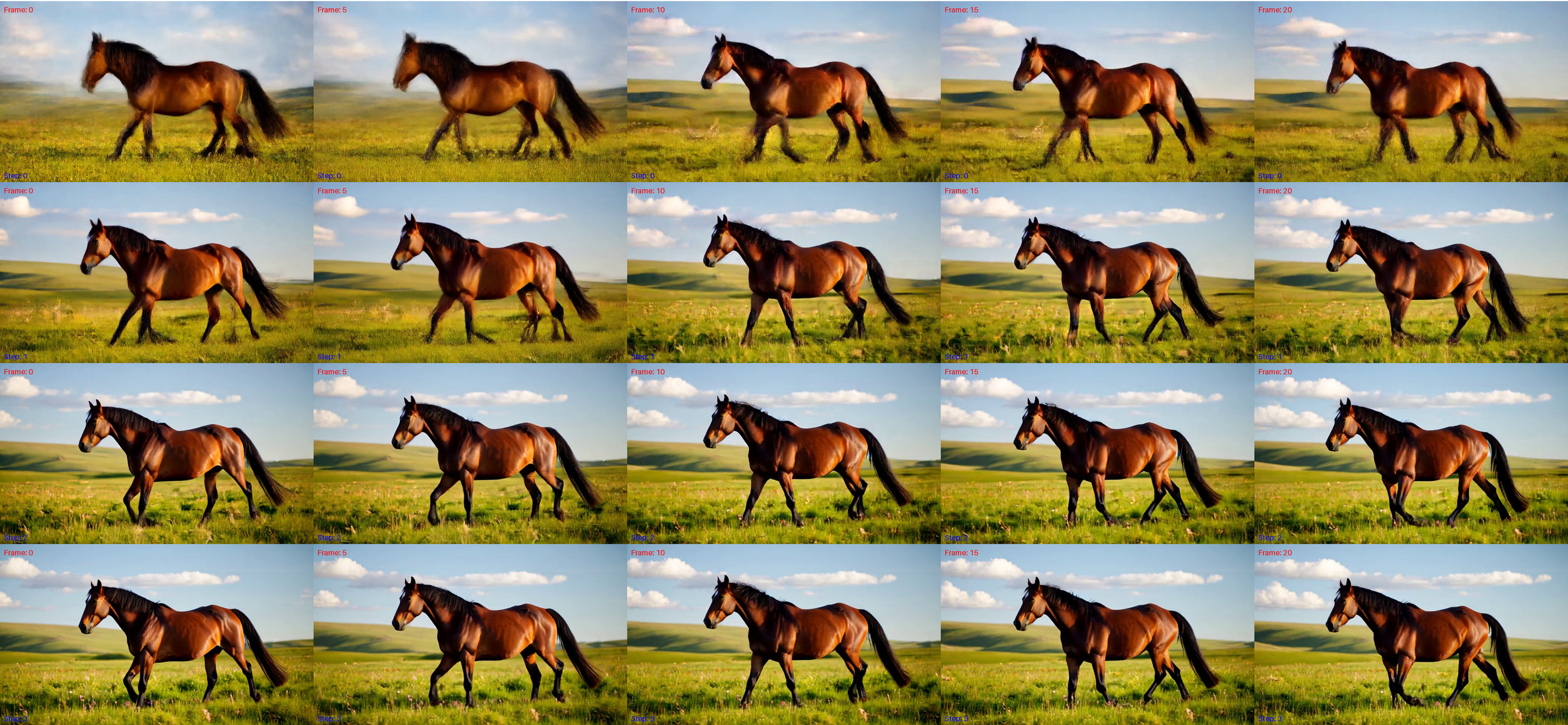}
\caption{\textbf{In-Context Forcing (Ours) Denoising Visualization.} The multi-step generation (first 20 frames, sampled every 5 frames) across sequential denoising steps (rows) and frames (columns) explicitly demonstrates our method's adaptive coarse-to-fine hierarchy. By utilizing contexts with higher noise levels in early stages, the model effectively prevents excessive replication of previous frames, thereby achieving superior semantic alignment and natural motion coherence. As generation progresses, it naturally transitions to fine-grained detail refinement, ensuring continuous visual improvement while maintaining robust inter-frame dynamics.}
\label{fig:supp_denoise_ours}
\end{figure*}

In Figs.~\ref{fig:supp_denoise_sf} and~\ref{fig:supp_denoise_ours}, we visualize the progressive denoising results of Self Forcing and our In-Context Forcing across different sampling steps (displaying the first 20 frames, sampled every 5 frames). The comparison reveals that Self Forcing, which strictly conditions on fully denoised contexts, forces an early restoration of fine-grained details. As shown in the top row of Fig.~\ref{fig:supp_denoise_sf}, the model achieves near-complete detail formation as early as the initial denoising step. While this appears to accelerate generation, it forces the model to directly replicate local details from preceding frames, severely compromising motion dynamics and leaving almost no capacity for meaningful refinement in subsequent steps. In contrast, In-Context Forcing leverages contexts with higher noise levels during the early stages. This provides adaptive guidance that prioritizes global semantic coherence and natural motion patterns over immediate detail synthesis. As generation progresses to later steps (subsequent rows in Fig.~\ref{fig:supp_denoise_ours}), the model smoothly transitions to fine-grained detail refinement. This behavior perfectly aligns with the fundamental coarse-to-fine hierarchy inherent in diffusion models, enabling continuous improvement in both visual fidelity and temporal dynamics throughout the entire generation process.

\section{User Study}
\label{sec:supp_user}
\textbf{User Preference Study.}
In the user preference study, we randomly selected extended prompts from VBench~\cite{huang2024vbench} and compared our method against three baselines: Wan2.1~\cite{wang2025wan}, CausVid~\cite{yin2025causvid}, and Self Forcing~\cite{huang2025selfforcing}. In each A/B test trial, participants evaluated video pairs randomly sampled from our method and each baseline. Results presented in Fig.~\ref{fig:user_study} demonstrate a consistent user preference for our approach across all comparisons.

\begin{figure*}[!t]
\centering
\includegraphics[width=\textwidth]{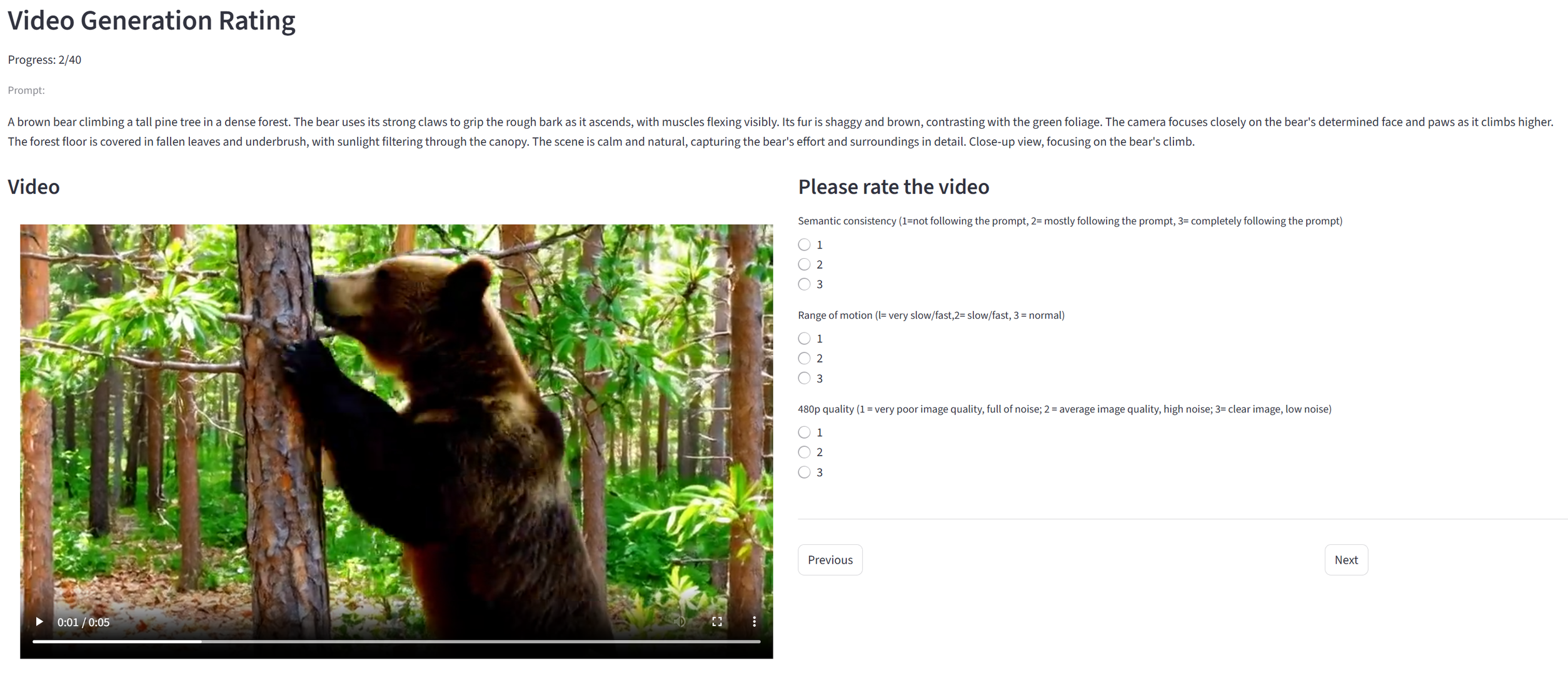}
\caption{\textbf{User Study Interface.} The interface presents the test video in the left panel and the evaluation controls, i.e., the three rating dimensions and the navigation button, in the right panel.}
\label{fig:supp_user_interface}
\end{figure*}

\textbf{Comprehensive Score Evaluation.}
For comprehensive score evaluation, each participant assessed randomly selected videos generated by our method and Self Forcing, providing ratings across three key dimensions: range of motion, visual quality, and semantic consistency. Fig.~\ref{fig:supp_user_interface} illustrates the user study interface, where the left panel displays the test video and the right panel presents the corresponding evaluation dimensions and rating buttons. After completing the ratings for all dimensions, the participants can proceed to the next video using the navigation button below.

\textbf{Results Analysis.}
As detailed in Table~\ref{tab:supp_user}, the comprehensive score evaluation specifically compares our method against the primary baseline, Self Forcing. Our In-Context Forcing consistently achieves higher ratings across all dimensions. Most notably, the substantial improvement in the motion metric ($2.73$ versus $2.31$) provides strong quantitative evidence that our progressive contextual modeling effectively prevents shortcut learning, resulting in vastly superior temporal dynamics. Furthermore, our approach yields clear enhancements in both global semantic alignment ($2.55$ versus $2.37$) and overall visual quality ($2.53$ versus $2.49$).

\begin{table}[!t]
\caption{Comprehensive Score Evaluation.}
\label{tab:supp_user}
\centering
\begin{tabular}{lccc}
\toprule
Method & Semantic & Motion & Quality \\
\midrule
Self Forcing & 2.37 & 2.31 & 2.49 \\
\textbf{In-Context Forcing (Ours)} & \textbf{2.55} & \textbf{2.73} & \textbf{2.53} \\
\bottomrule
\end{tabular}
\end{table}

\section{Limitations and Future Directions}
\label{sec:supp_limit}
\textbf{Limitations.}
A primary limitation of our approach lies in an additional training convergence time. Because our method introduces progressive contexts with decreasing noise levels, the model requires more training iterations to fully adapt to these diverse contextual signals compared to the baseline. Specifically, under identical experimental settings, the Self Forcing baseline reaches optimal performance at $1500$ iterations, whereas our method requires $1900$ iterations. However, this computational overhead is strictly confined to the training phase. We consider this additional training duration a justified investment for the significantly enhanced semantic alignment, robust temporal dynamics, and highly efficient inference achieved during deployment.

\textbf{Future Directions.}
While this work primarily focuses on exploring the impact of progressive contexts in few-step autoregressive generation, our attention-map visualizations (Fig.~\ref{fig:supp_attn}) reveal an interesting phenomenon: the attention patterns remain inherently sparse even after autoregressive distillation. This persistent sparsity suggests significant potential for further structural optimization. Therefore, a highly promising direction for future research is to investigate the integration of attention sparsification techniques with few-step autoregressive models. Leveraging this inherent sparsity could substantially reduce the computational overhead and memory requirements for per-frame attention calculations, thereby amplifying the benefits of parallel denoising and enabling even greater scalability.

\section{Additional Qualitative Results}
\label{sec:supp_qual}
\begin{figure*}[!t]
\centering
\includegraphics[width=\textwidth]{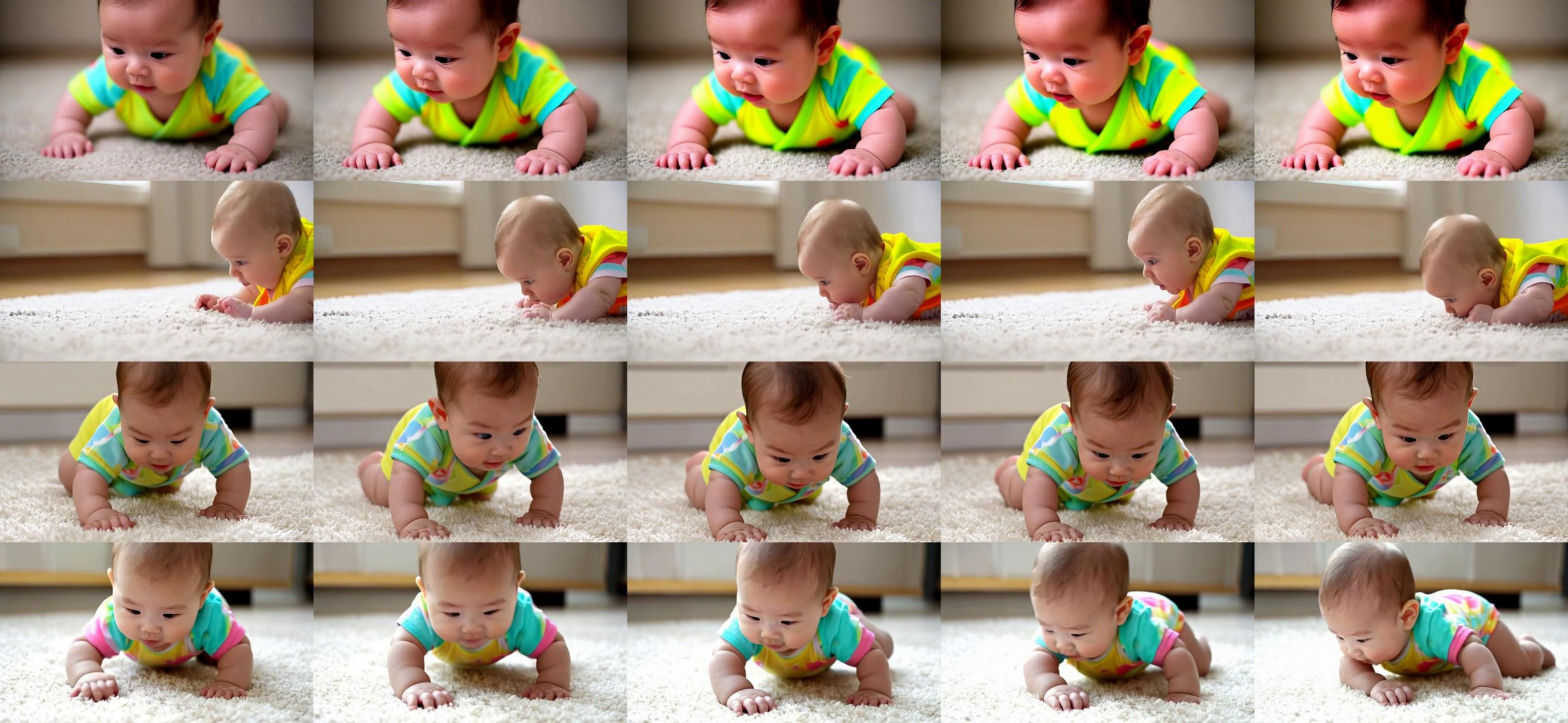}
\caption{Additional qualitative comparisons of short-video generation on VBench. Rows from top to bottom: CausVid, Wan2.1, Self Forcing, and In-Context Forcing (Ours). \textit{All visual results are generated through text-to-video (T2V) inference} using the prompt: ``A close-up of a baby crawling on a soft, carpeted floor. The baby is on all fours, with chubby arms and legs pushing and pulling themselves forward. They have a curious expression, with their head tilted upwards as they explore their surroundings. The baby is dressed in a bright, colorful onesie. The camera remains static, focusing solely on the baby's movements and facial expressions, capturing each small step and moment of discovery.''}
\label{fig:supp_short_baby}
\end{figure*}

\begin{figure*}[!t]
\centering
\includegraphics[width=\textwidth]{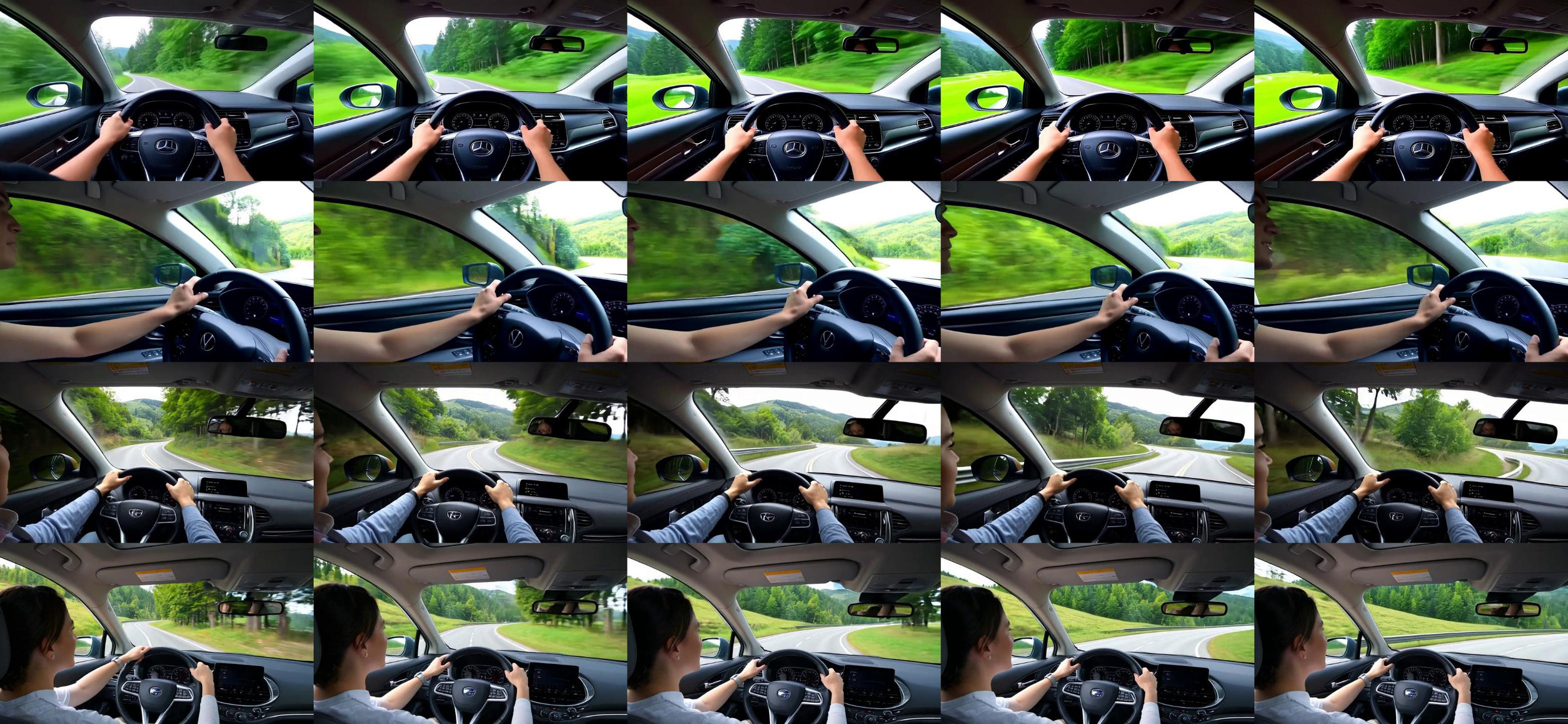}
\caption{Additional qualitative comparisons of short-video generation on VBench. Rows from top to bottom: CausVid, Wan2.1, Self Forcing, and In-Context Forcing (Ours). \textit{All visual results are generated through text-to-video (T2V) inference} using the prompt: ``A person is driving a modern sedan on a winding road surrounded by lush greenery. The driver is focused, with a neutral expression, and their hands are positioned at ten and two o'clock on the steering wheel. The car is moving smoothly along the curving path, and the scenery outside the window changes as the vehicle progresses. The background includes tall trees and rolling hills, creating a serene driving environment. The scene is captured from inside the car, providing a first-person perspective of the driver and the road ahead. Medium shot focusing on the driver and the immediate surroundings of the car.''}
\label{fig:supp_short_driving}
\end{figure*}

\begin{figure*}[!t]
\centering
\includegraphics[width=\textwidth]{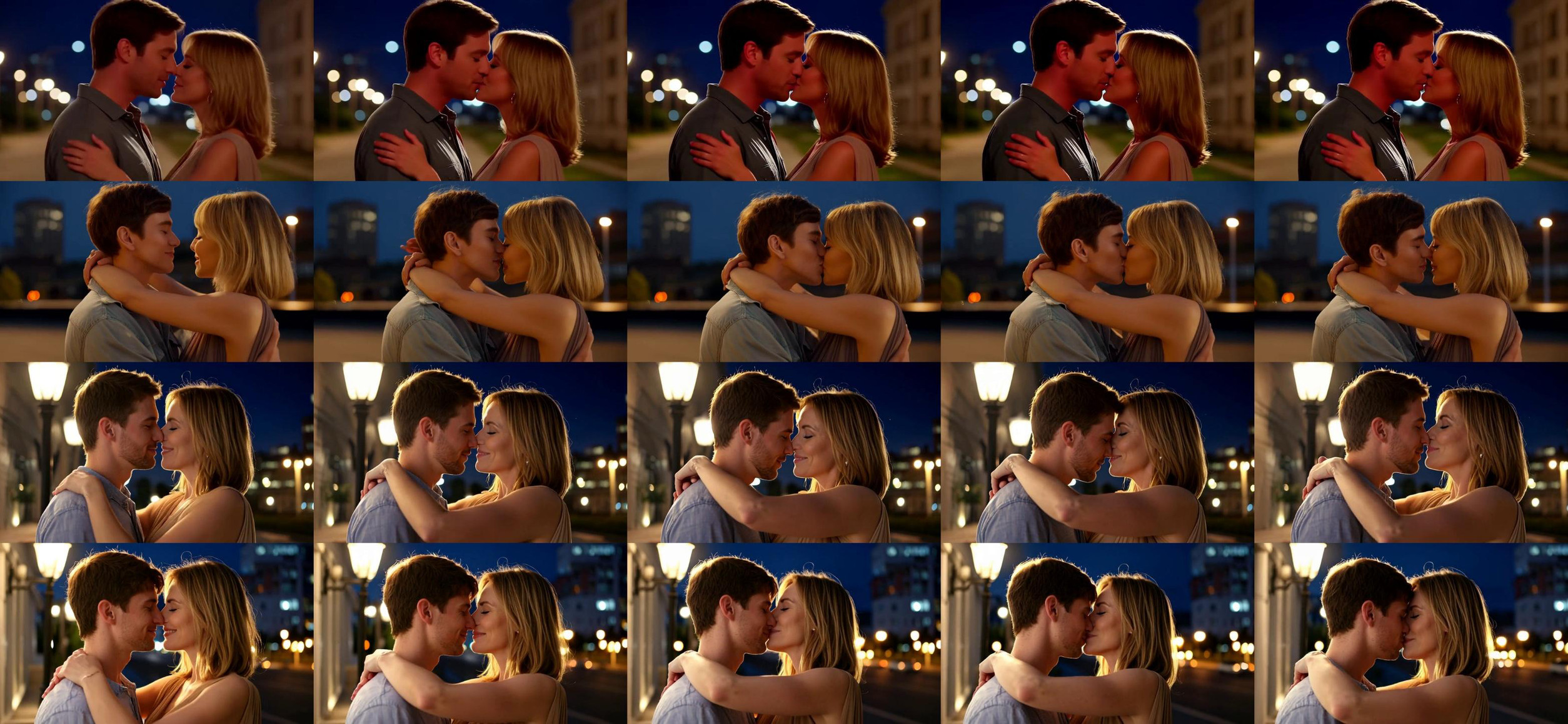}
\caption{Additional qualitative comparisons of short-video generation on VBench. Rows from top to bottom: CausVid, Wan2.1, Self Forcing, and In-Context Forcing (Ours). \textit{All visual results are generated through text-to-video (T2V) inference} using the prompt: ``A romantic close-up of two people kissing passionately. They are standing outdoors under a softly lit streetlamp at night. Both individuals have their eyes closed, leaning into each other with gentle expressions of affection. The man has short brown hair and is wearing a casual shirt, while the woman has shoulder-length blonde hair and is dressed in a flowy evening gown. Their arms are wrapped around each other, pulling them closer together. The background shows blurred city lights and buildings, adding to the intimate atmosphere. The scene captures the moment of deep connection between them, with a soft focus and warm lighting.''}
\label{fig:supp_short_kissing}
\end{figure*}

\begin{figure*}[!t]
\centering
\includegraphics[width=\textwidth]{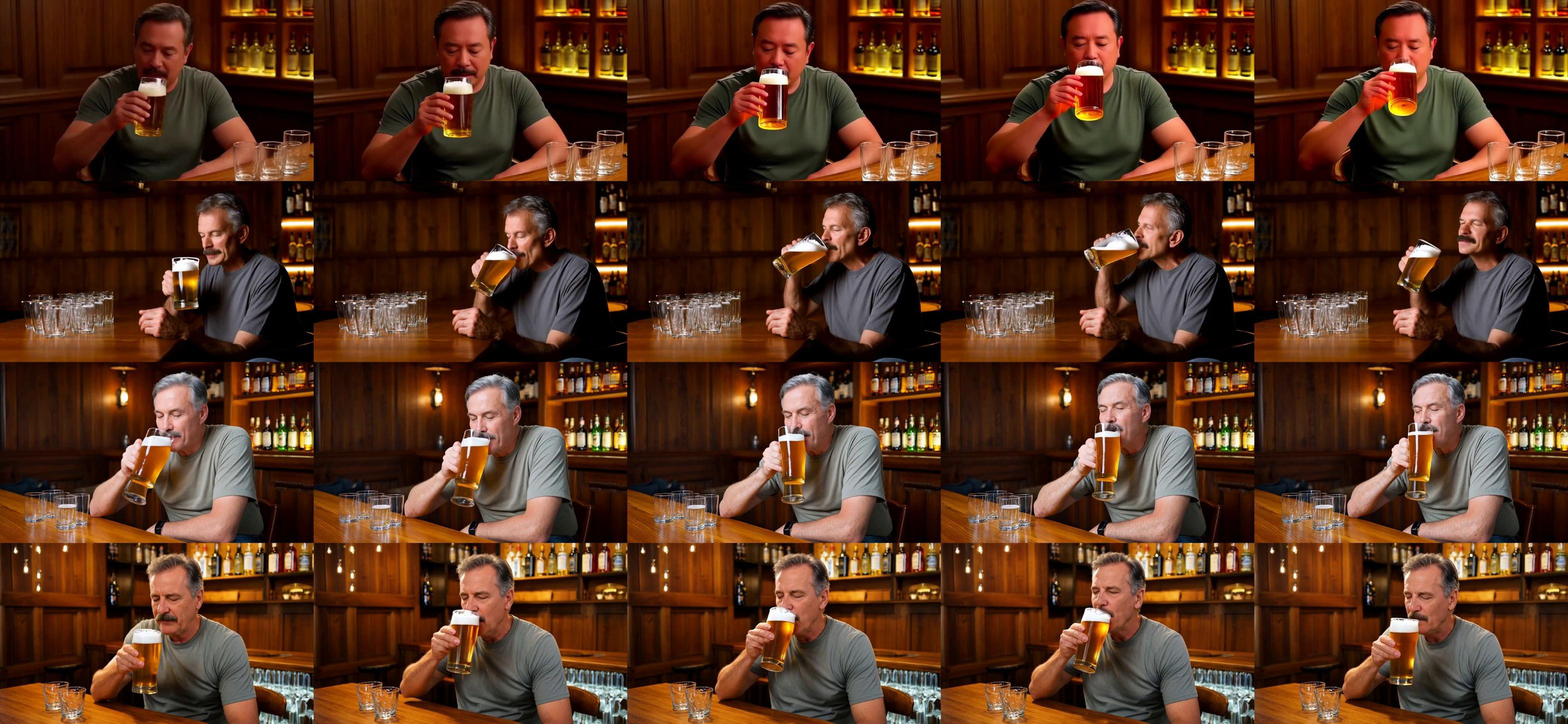}
\caption{Additional qualitative comparisons of short-video generation on VBench. Rows from top to bottom: CausVid, Wan2.1, Self Forcing, and In-Context Forcing (Ours). \textit{All visual results are generated through text-to-video (T2V) inference} using the prompt: ``A middle-aged man with a casual outfit, including a t-shirt and jeans, is tasting a frothy beer from a pint glass. He has a mustache and is sitting at a wooden bar table with several empty glasses nearby. His face shows a thoughtful expression as he savors the taste, tilting his head slightly and closing his eyes. The bar has warm, ambient lighting and rustic decor, with wooden panels and dimly lit bottles of liquor on shelves behind him. Medium close-up shot focusing on his face and the beer glass.''}
\label{fig:supp_short_beer}
\end{figure*}

\begin{figure*}[!t]
\centering
\includegraphics[width=\textwidth]{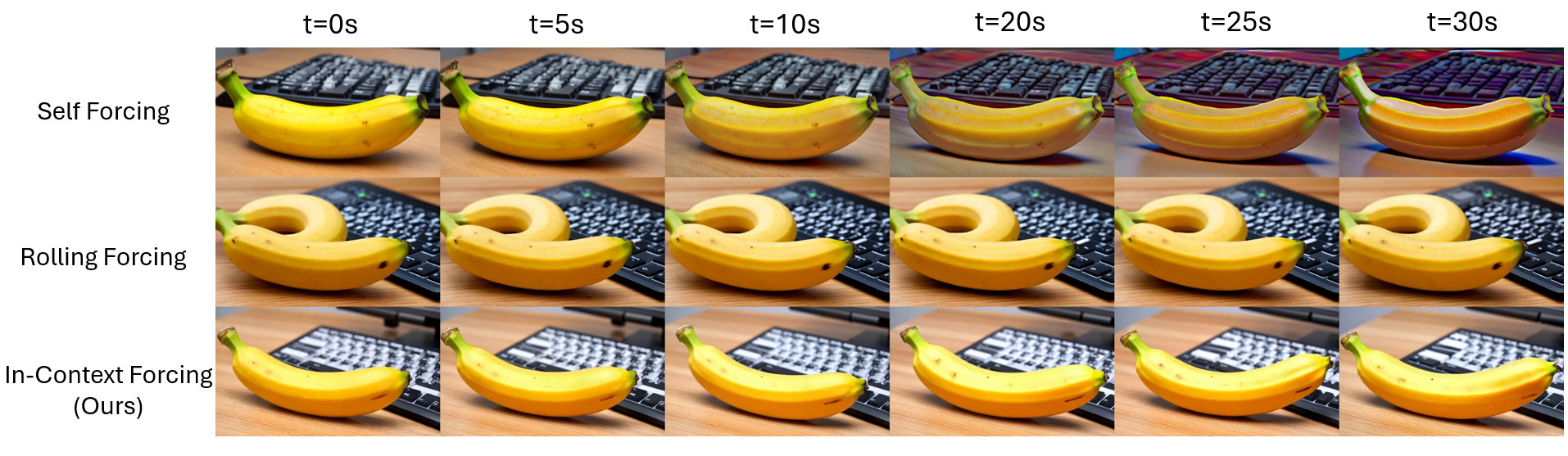}
\caption{Additional qualitative comparisons of 30-second long-video generation on VBench. \textit{All visual results are generated through text-to-video (T2V) inference} using the prompt: ``A majestic great white shark is swimming gracefully through the vast ocean in a watercolor painting style. The shark's sleek body is painted with shades of grey and white, blending smoothly with the surrounding water. The water is depicted with soft blues and greens, showing gentle waves and sunlight filtering through, creating a serene underwater atmosphere. Schools of smaller fish swim alongside the shark, adding to the vibrant marine life. The background showcases a distant coral reef and floating seaweed, enhancing the sense of depth and life in the ocean. The painting captures the shark in a mid-swim pose, with its powerful tail propelling it forward. Medium shot, focusing on the shark and immediate surroundings.''}
\label{fig:supp_long_shark}
\end{figure*}

\begin{figure*}[!t]
\centering
\includegraphics[width=\textwidth]{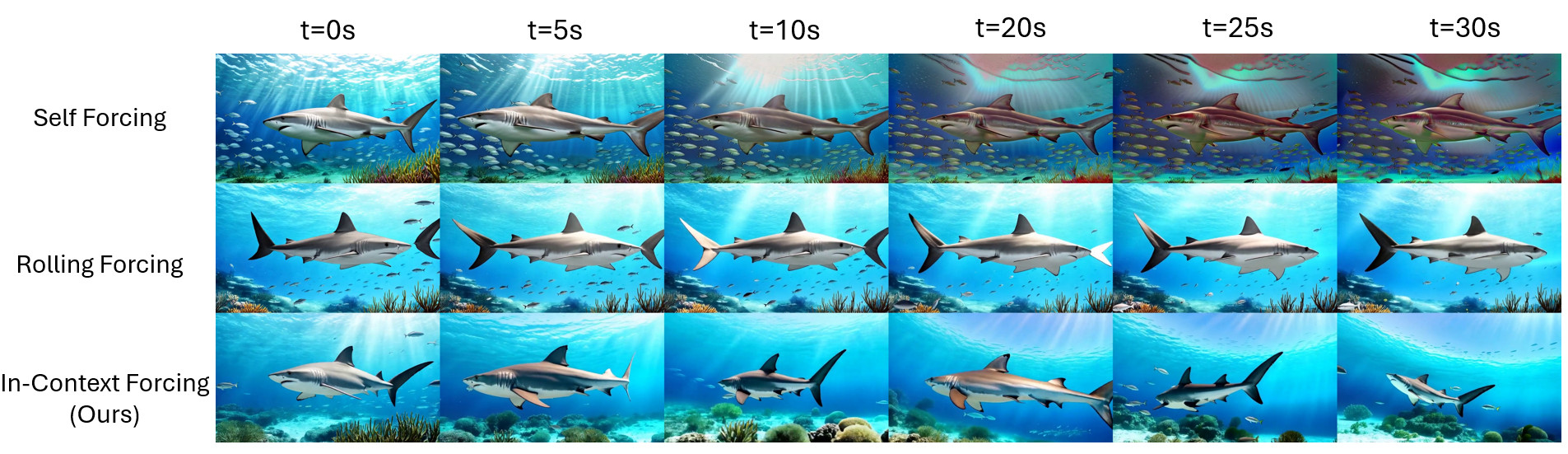}
\caption{Additional qualitative comparisons of 30-second long-video generation on VBench. \textit{All visual results are generated through text-to-video (T2V) inference} using the prompt: ``A close-up view of a ripe yellow banana and a black keyboard lying side by side on a wooden table. The banana is curved with a few spots, indicating it is just right for eating. The keyboard has a sleek design with white keys and black letters. The banana is positioned closer to the front of the frame, while the keyboard is slightly behind it, creating depth. The background is blurred, focusing attention on these two items. The banana appears fresh and inviting, contrasting with the utilitarian nature of the keyboard. Static scene, no camera movement.''}
\label{fig:supp_long_banana}
\end{figure*}

\begin{figure*}[!t]
\centering
\includegraphics[width=0.75\textwidth]{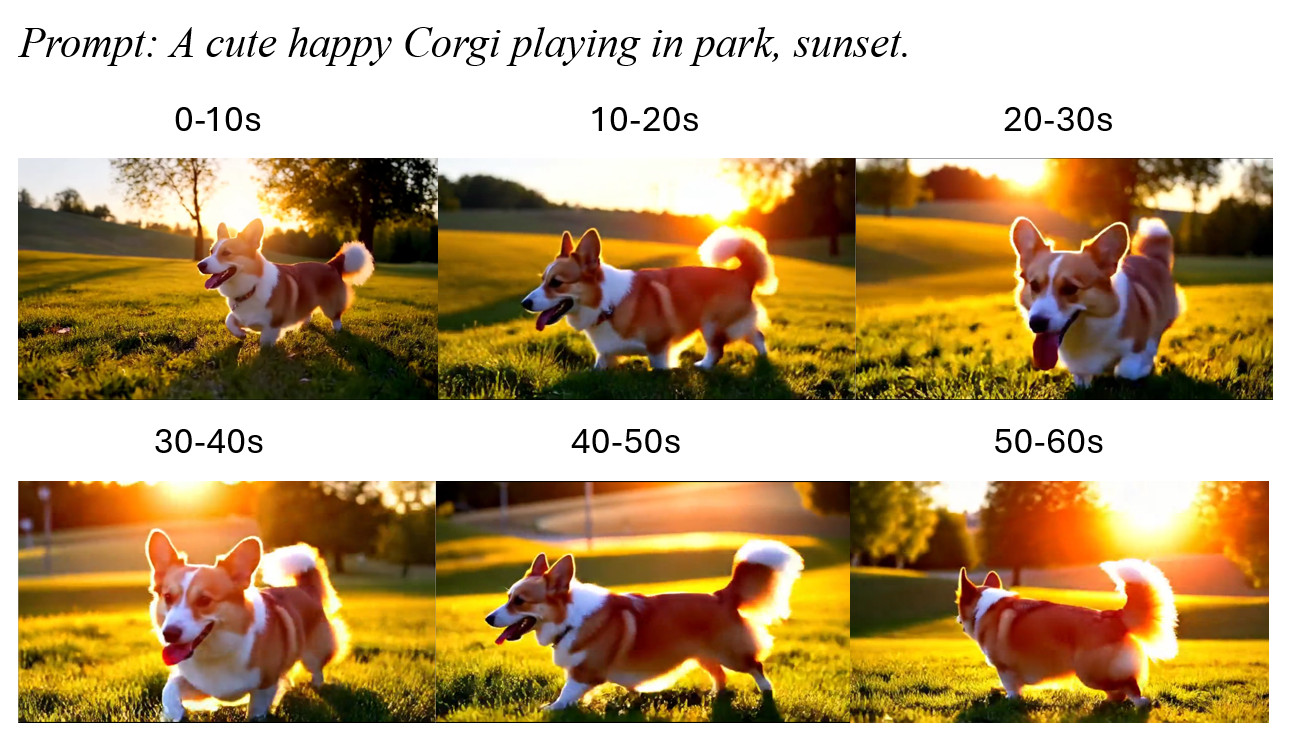}
\caption{Qualitative results of In-Context Forcing for 60-second long-video generation on VBench. \textit{The visual results are generated through text-to-video (T2V) inference} using the prompt: ``A cute happy Corgi playing in park, sunset.''}
\label{fig:supp_long_60s}
\end{figure*}

We present additional qualitative results to demonstrate the advantages of In-Context Forcing in visual quality and temporal coherence across varying video lengths. These results complement the comparisons already reported in the main paper, i.e., Fig.~\ref{fig:short_qualitative} for short clips and Fig.~\ref{fig:long_qualitative} for 30-second sequences. Specifically, Figs.~\ref{fig:supp_short_baby} to~\ref{fig:supp_short_beer} show short-video comparisons on VBench against the baselines, while Figs.~\ref{fig:supp_long_shark} and~\ref{fig:supp_long_banana} extend these comparisons to 30-second scenarios. Finally, Fig.~\ref{fig:supp_long_60s} highlights our method's scalability with a 60-second video. Across all cases, In-Context Forcing sustains richer motion and stronger semantic alignment with the prompt, whereas the baselines tend to replicate patterns from preceding frames and gradually lose temporal dynamics.

\end{document}